\documentclass[11pt, a4paper, logo, internal]{dm}
\usepackage{graphicx}%
\usepackage{multirow}%
\usepackage{amsmath,amssymb,amsfonts}%
\usepackage{amsthm}%
\usepackage{mathrsfs}%
\usepackage[title]{appendix}%
\usepackage{xcolor}%
\usepackage{textcomp}%
\usepackage{manyfoot}%
\usepackage{booktabs}%
\usepackage{algorithm}%
\usepackage{algorithmicx}%
\usepackage{algpseudocode}%
\usepackage{listings}%
\usepackage{hyperref}

\usepackage[commandnameprefix=always]{changes}
\usepackage[most]{tcolorbox}
      
\def\method{LabAgent}

\theoremstyle{thmstyleone}%
\theoremstyle{thmstyletwo}%

\theoremstyle{thmstylethree}%
\usepackage{mathtools}
\usepackage[dvipsnames]{xcolor}
\usepackage[numbers, sort&compress, round]{natbib}
\usepackage{booktabs}
\usepackage{graphicx}
\usepackage{xfrac}
\usepackage{bbm}
\usepackage{changes}
\usepackage{rotating}
\usepackage{wrapfig}

\usepackage[most]{tcolorbox}
\usepackage{xparse}
\usepackage{adjustbox}
\usepackage{xspace}
\usepackage{changepage}
\usepackage{enumitem}
\usepackage{pifont}
\usepackage{ulem}
\usepackage{tocloft}
\usepackage[toc]{multitoc}
\usepackage{etoc}
\usepackage{dsfont}
\usepackage{dm-colors}
\usepackage{multirow}
\usepackage{minted}
\usepackage{caption}
\usepackage{soul}
\usepackage{float}
\usepackage{svg}
\usepackage{adjustbox}
\usepackage{setspace}
\svgsetup{inkscapelatex=false}
\usepackage{silence}  
\newcommand{\hide}[1]{}

\pdftrailerid{redacted}

\title{LabAgent: Customize Any Research Hubs for Scientific Discoveries Using AI Agents}

\begin{document}

\author[1]{Lei Liu}
\author[2,3]{Yikun Zhang}
\author[4]{Jialin Chen}
\author[5]{Wanjia Zhao}
\author[4]{Rex Ying}
\author[2,3]{Wengong Jin}
\author[6,7]{Hua Xu}
\author[8]{James Zou}
\author[1,3,6,7]{Tianyu Liu*}
\author[1,6]{Hongyu Zhao*}

\affil[1]{Department of Biostatistics, Yale University}
\affil[2]{Department of Computer Science, Northeastern University}
\affil[3]{EWSC Center, Broad Institute of MIT and Harvard}
\affil[4]{Department of Computer Science, Yale University}
\affil[5]{Department of Computer Science, Stanford University}
\affil[6]{Interdepartmental Program in Computational Biology and Bioinformatics, Yale University}
\affil[7]{Department of Biomedical Informatics and Data Science, Yale University}
\affil[8]{Department of Biomedical Data Science, Stanford University}

\affil[*]{Corresponding authors. Contact emails:
tianyu.liu\@yale.edu; hongyu.zhao\@yale.edu}




\begin{abstract}
Scientific research is a continuous process that emphasizes inheritance. Methods developed by predecessors are often expanded upon by new researchers to explore more novel and in-depth scientific questions. However, the change of lab staff, such as student graduation, leads to a lack of personnel capable of replicating methods. Methods that have been developed with significant effort and resources cannot be continued. To address these limitations, we propose \method{}, a reproduce and discovery harness tailored for a lab's continuous work. \method{} employs two mechanisms to guarantee that all skills can be executed and verified and to record the corrective methods and experiences, allowing for direct correction or avoidance of similar errors. We applied \method{} to drug property prediction, biomedical problem analysis, protein variant effect prediction, and statistical genetics in life science domains. \method{} ranks first over commercial generalist agents in every domain, and demonstrates accurate reproduction of a published figure. Overall, these results demonstrate that \method{} can effectively integrate and reasonably expand laboratory knowledge.
\end{abstract}




\maketitle
\section{Introduction}
Computational methods have become central to modern scientific
discovery, especially in biology\cite{wang2023scientific,wei2025aiscienceagenticscience}. Across single-cell
transcriptomic analysis\cite{Cui2024scGPTTB,Liu2025UNICORNTU}, drug property
prediction\cite{Huang2021TherapeuticsDC,Liu2025BuildingAU}, protein variant effect
prediction\cite{notin2023proteingym}, and genetic association mapping\cite{Wang2018ASN},
researchers have built a vast array of models, algorithms, and data-processing
pipelines and shared them openly through public code repositories, tutorials,
and datasets\cite{Wilkinson2016TheFG}. This culture of open sharing has accelerated
biological research. A method developed in one laboratory can be reproduced,
reused, and extended by peers worldwide, letting the field build cumulatively
on prior work rather than starting each study from scratch. In practice,
though, much of this shared work is hard to reuse. Codes can be
public and its paper carefully written, yet reproducing the reported result
often fails\cite{Baker20161500SL}, because running a method correctly depends
on far more than the code that is released. What the code leaves out is a body
of implicit knowledge, such as a preprocessing step that no one wrote down, a
package version that was never pinned, or a parameter chosen by hand for the
dataset it was tuned on. Because this knowledge is specific to the laboratory
that produced it, it survives only tacitly, in the fine details of a
repository, in scattered issue-tracker discussions, and in the memory of
individual researchers\cite{li2026papersdonttellyou}. As lab members move
on, it gradually erodes, taking with it the ability to build on what the
laboratory actually did. A different goal is therefore to build a system that
recovers this knowledge directly from a laboratory's public artifacts,
including its papers, code, tutorials, and issue discussions. Such a system
would let others reliably run and verify the laboratory's methods, and build
on these validated methods to make new discoveries.

Large Language Model (LLM)-driven agents possibly can provide such a system.
Agents built on LLMs can read code, call tools, execute programs, and query
external knowledge, and a growing number have been applied across the
scientific workflow. But each was built for a different goal, and none
provides more than a fragment of what such a system would need. Discovery
agents, such as Google's AI co-scientist, reason from the existing literature
toward new hypotheses, and can propose ideas that later hold up at the
bench\cite{gottweis2026accelerating,Boiko2023AutonomousCR,swanson2025virtual,lu2026towards,Mitchener2025KosmosAA,roohani2025biodiscoveryagent,wang2025geneagent,jin2025stellaselfevolvingllmagent,Huang2025.05.30.656746}.
Engineering agents, such as SWE-agent, understand and execute code well enough
to resolve real issues in software
repositories\cite{yang2024swe,wang2025openhands,zhao2026autoreproduce,nam2026mle,jiang2025aideaidrivenexplorationspace,yang2025rdagentllmagentframeworkautonomous}.
Benchmarks, such as PaperBench, confirm that these abilities can be
measured\cite{pmlr-v267-starace25a,ICLR2025_7e3767db,mitchener2025bixbenchcomprehensivebenchmarkllmbased}.  Biomedical work sets demanding benchmarks of its own. They ask whether an
agent understands an analysis and can assemble the workflow that answers it.
BiomniBench-DA rebuilds the analyses of published papers across a range of
disease areas and scores the whole trajectory against an expert's
rubric\cite{Qu2026.05.12.724604}. GeneBench-Pro poses multi-stage problems
across genetics and omics and scores only the estimate at the
end\cite{Li2026.06.29.735386}. Each of these captures something
such a system would need, yet none is organized around a specific laboratory's validated
methods and the procedures that run them. Discovery agents draw on the general literature and public
databases, not the methods a particular laboratory has developed and verified.
Engineering agents handle each task in isolation and do not turn what they
learn into knowledge the laboratory can reuse, or carry it into discovery. The
most capable of them, the AI co-scientist, runs specialized agents through a
generate, debate, and evolve loop over the published literature and has
produced hypotheses confirmed experimentally\cite{gottweis2026accelerating}. Yet even it
neither reproduces nor verifies any laboratory's methods, and the hypotheses
it proposes must still be implemented and tested by others. Assembling these
fragments therefore takes more than a stronger agent. It requires a way to
package the knowledge behind a laboratory's methods so that others can reuse
it, rather than rebuild it for each new task.

Skills are one of the key factors in helping agents complete reproduction tasks. A skill packages a task's procedural
knowledge, down to the details a method usually leaves unstated, into a
reusable unit that an agent can invoke on demand\cite{claude_agent_skills}. In this
form the knowledge becomes explicit and executable, so it persists even as a
laboratory's members move on. Building such skills by hand does not scale,
because a laboratory's know-how is spread across far more material than anyone
can package manually. Recent work has therefore begun to build them
automatically, mining skills from heterogeneous scientific resources such as
code, documentation, and papers, validating and repairing them by execution,
and organizing them into self-evolving libraries that improve agent
performance on scientific tasks\cite{shen2026skillfoundrybuildingselfevolvingagent,huang2026cascadecumulativeagenticskill,zheng2025skillweaverwebagentsselfimprove,Wang2023VoyagerAO}.
This line of work, however, builds a skill library for a field as a whole,
drawn from shared community resources rather than from one laboratory's own
papers, code, and issue discussions. Its skills, moreover, are validated for
executability and general utility rather than against the results a specific
study reports, so such systems do not reproduce a target method and confirm
that the reproduction matches the original. Nor do these libraries retain the
errors and fixes from reproducing one method as experience that guides the
next, so each reproduction begins again rather than building on the last.
Finally, they stop at improving task performance, without turning a
laboratory's validated methods toward proposing and testing the new hypotheses
that reuse is meant to enable.

Here we present \emph{LabAgent}, a multi-agent framework that instantiates a laboratory-specific
research agent directly from a laboratory's public artifacts. An explore agent
reads the laboratory's papers, code repositories, tutorials, and issue
discussions, grounds the methods described in them against the code that
implements them, and synthesizes the results into executable skills. Each skill
records not only what a method does but how it must be run, including the
preprocessing steps a paper omits, the package versions a repository pins, and
the failure modes its users have already reported. The resulting library
belongs to that laboratory rather than to a field as a whole, and captures the
methods, conventions, and ongoing directions of the laboratory that produced
it. A reproduce agent then puts these skills to work. Given a new reproduction
target, it retrieves protocols and fixes from earlier reproductions and
assembles a set of role-specialized agents to plan, implement, run, and verify
the method. It then compares what it obtains against the metrics the original
study reported. Verification is therefore against the science, not merely
against the absence of errors. A run that completes without crashing but does
not recover the published numbers is treated as a failure. When a run does
fail, the agent localizes the failure to a specific skill and to the step at
which that skill broke, repairs the skill, and retries. The execution logs,
diagnoses, and fixes accumulated along the way are written back to a persistent
memory, so that each new reproduction is conditioned on what earlier ones
learned rather than starting from scratch. Building on the skills that survive
this verification, \method{} turns the laboratory's own validated methods toward
proposing and testing new scientific hypotheses. LabAgent thus moves scientific
agents from task-centric automation toward reconstructing a laboratory's
research program.

\section{Results}
\begin{figure}[tp]
\centering
\includegraphics[width=0.9\linewidth]{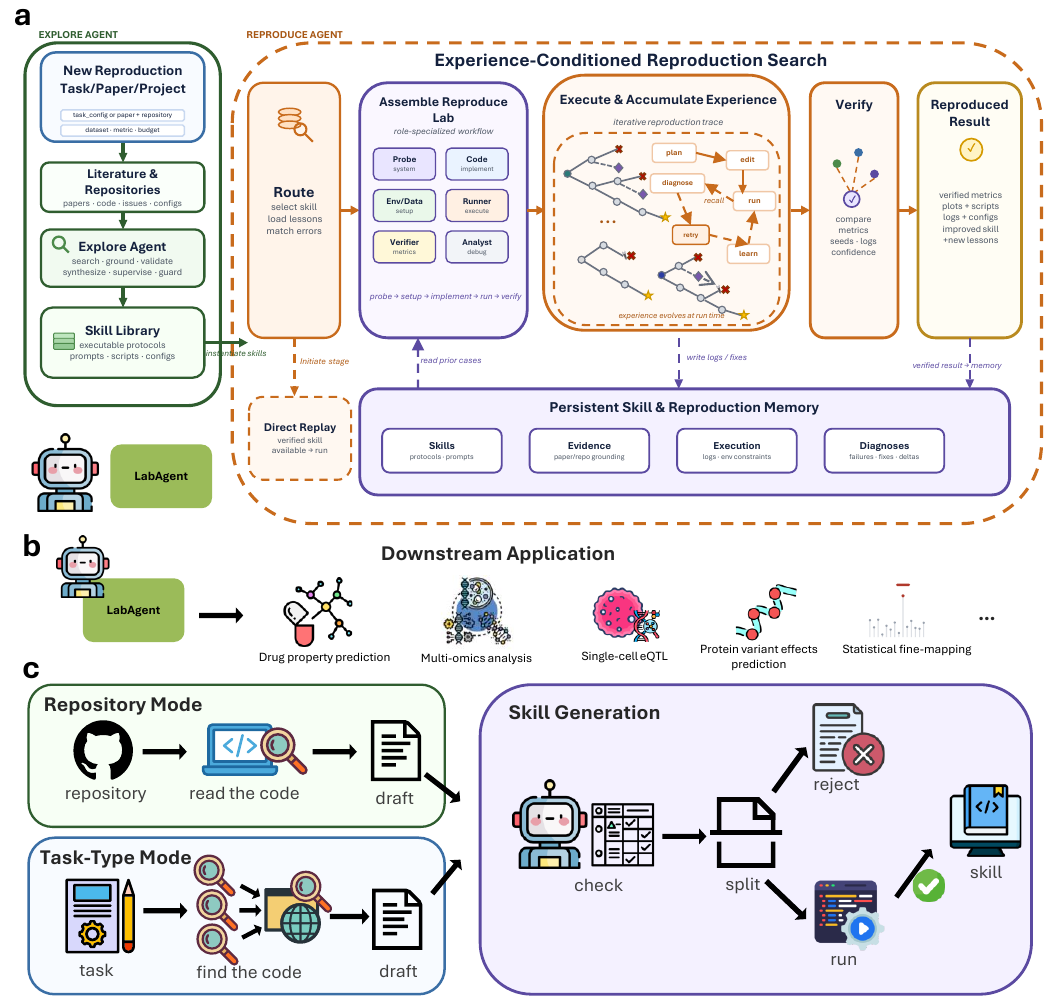}
\caption{\textbf{The \method{} framework.}
(a) The framework of \method{}. A target reaches the explore agent as a task configuration or as a paper with its repository, dataset, metric and budget. That agent searches the literature and code and deposits an executable skill in the library. The reproduce agent routes the target to one skill, loads its lessons, and works it through six roles from a machine probe to a failure analysis. It compares the result with the published value and returns verified metrics with the scripts, logs and configurations behind them, and an improved skill. The dashed path replays a verified skill. A persistent memory holds the skills, their evidence, the logs and the diagnoses that pair a failure with its fix, and every stage reads and writes it.
(b) Downstream applications. Five settings of the Results, from drug property prediction on public leaderboards to fine-mapping with no released figure code.
(c) How a skill is made. The agent reads a repository when the target names one and searches for the code when it does not. Both routes reach one draft, and a second model call checks it against the rubrics. Only a skill that runs end to end enters the library.}
\label{fig:framework}
\end{figure}

\noindent\textbf{Overview of \method{}.}
We present three contributions of \method{} as a harness for reproduction and
discovery. First, we build a route from a laboratory's public code repository
to an executable skill, and we admit a skill to the library only after it has
run end to end. A library we assemble this way lets a hand-built agent land
inside the published error bar more often than either commercial harness we
compare against. Second, we carry the experience of each run into a memory
that later runs read back. A run leaves behind the command that cleared each
failure it met, and a later run opens with those commands already in hand. By
contrast, a commercial harness leaves no record of that kind. Third, we work
both mechanisms through four domains of life science, and the four settings
differ in what the published record withholds. On drug property prediction the
method is named and its code is public, and \method{} rebuilds the leaderboards
of the Therapeutics Data Commons from what each entry released. On genomic and
single-cell analysis no protocol exists, and \method{} carries an open analysis
through a chain of dependent decisions and recovers a nuisance variable that
the task never names. On protein variant effect prediction no method is named,
and \method{} chooses among dozens of published predictors on its own. On
statistical genetics no figure code exists, and \method{} rebuilds the
experiment behind a published figure and leaves both of its departures from
that figure locatable and named. Based on this design, \method{} is able not only to reproduce the selected methods but also to search for combinations of different solutions and produce discoveries. Our comparison versus state-of-the-art agentic harness also strengthens our conclusion.
We evaluate the framework on three benchmarks and two case studies.
Fig.~\ref{fig:framework}a lays out the framework, and we describe both agents
and the memory in the Methods section.

\textbf{Reproducing drug property leaderboards from laboratory code.}
We first took a setting in which the right answer is already on record. The absorption, distribution, metabolism, excretion and toxicity (ADMET) leaderboards of the Therapeutics Data Commons rank published methods on 22 datasets, and every entry publishes both a value and the spread of that value. Those 22 datasets follow a compound through the body. Six of them ask how much
of it crosses the gut wall and reaches the blood, three ask where it travels
once it is there, six ask which cytochrome enzymes break it down, three ask
how fast it leaves, and four ask what harm it does along the way. An analysis
of the oral compounds that four large pharmaceutical companies carried into
development between 2000 and 2010 ties the physical properties of a compound
to its failure on safety grounds\cite{waring2015analysis}. A reproduction of the code
these studies release lets a later group build on that work and take its own
life-science question further. We took the top three entries of each
leaderboard and reached 65 of the 66 because one repository has been deleted upstream. Those 65 entries resolve to 11 distinct repositories, and the
explore agent built one skill from each. These entries are known to resist
rebuilding, and an independent assessment of the ten distinct models that
occupy the top three places found that only three could be
reproduced\cite{Koleiev2026CriticalAO}. We compared against Claude Code CLI and against Codex CLI. Claude Code runs on
Opus 4.8 and Codex runs on GPT-5.5.

\begin{figure}[tp]
\centering
\includegraphics[width=1\linewidth]{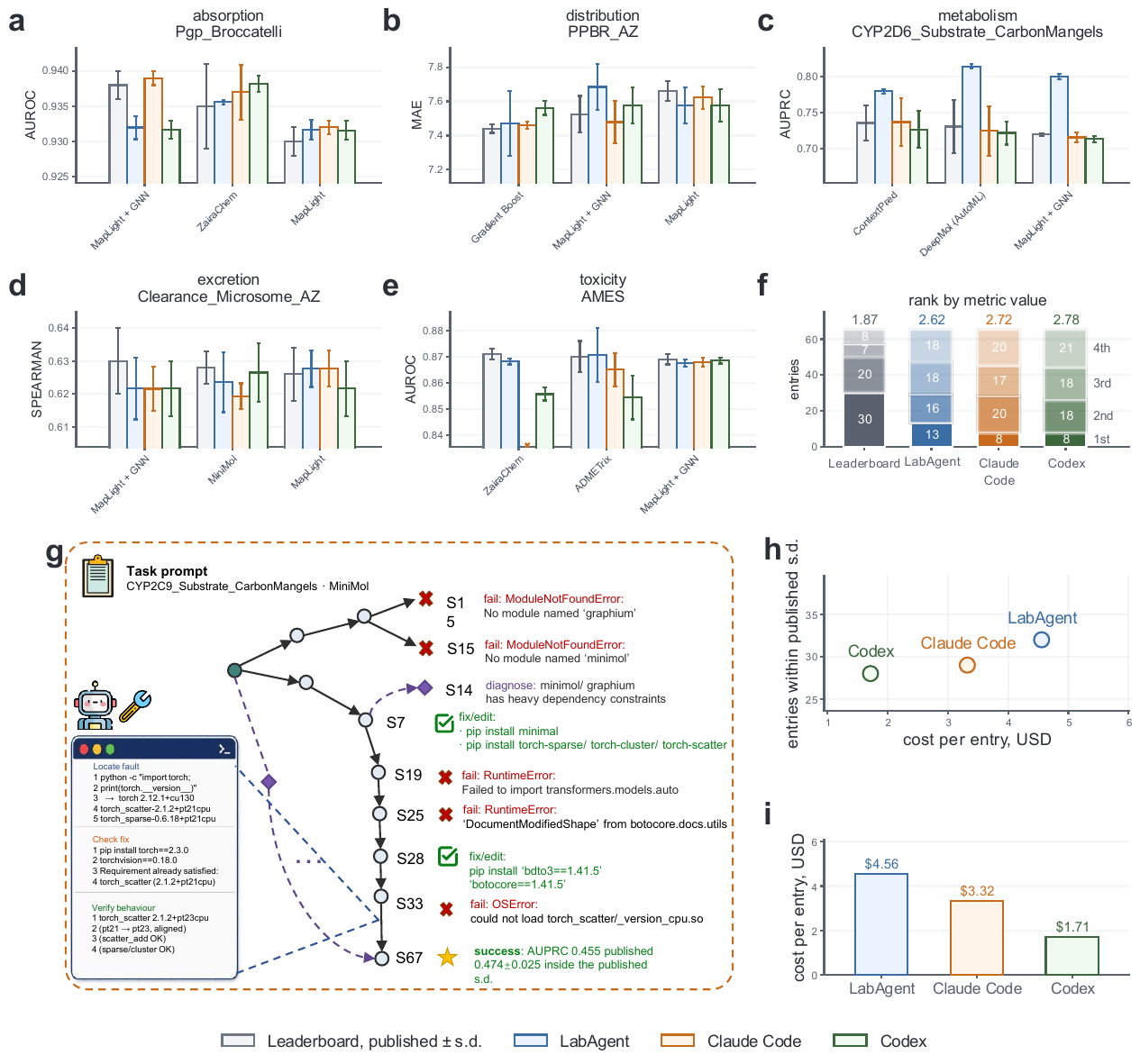}
\caption{\textbf{Reproducing published ADMET leaderboard values.}
(a)-(e) One representative dataset from each ADMET category,
with the three top-ranked entries of that leaderboard. Grey bars give the
published value and its standard deviation, and the coloured bars give what
\method{}, Claude Code and Codex obtained.
(a) Absorption. (b) Distribution. (c) Metabolism.
(d) Excretion. (e) Toxicity.
(f) Rank of each of the four values on all 65 entries, with the count at each rank and the mean rank above the bar.
(g) One reproduction trajectory over 67 steps, with the failures the agent met, the diagnoses it made and the fixes it applied. The run ends at an AUPRC of 0.455 against a published 0.474 $\pm$ 0.025.
(h) Entries that fall inside the published standard deviation against cost per entry.
(i) Cost per entry in US dollars.}
\label{fig:tdc}
\end{figure}

We compared the three agents from two angles. The first angle is performance, and we read it two ways. We counted the
entries that land inside the published error bar. \method{} landed inside the
published standard deviation on 32 of the 65 entries and on more of them than
either commercial harness (Fig.~\ref{fig:tdc}a--e). We then ranked the four
values available for every entry. \method{} took the best mean rank of the three
at 2.62, and it ranked first on 13 of the 65 entries against 8 for each of the
others (Fig.~\ref{fig:tdc}f). No agent matches the leaderboard itself at a mean
rank of 1.87. \method{} sits closest to it on both readings. The second angle is cost. \method{} spent US\$4.56 per entry, about a third more
than the closest harness (Fig.~\ref{fig:tdc}i). The two angles meet on one pair
of axes. \method{} reaches the highest count of entries inside the error bar and
pays the most for that count (Fig.~\ref{fig:tdc}h).

We traced one run end to end to show where the spend goes (Fig.~\ref{fig:tdc}g). In that run, the agent was asked to rebuild MiniMol on a substrate dataset, and its first two attempts failed because neither \texttt{graphium} nor \texttt{minimol} was importable. It diagnosed the dependency constraint at step 14, installed the
geometric extensions on a second branch, and then met a runtime error inside
the transformer loader and a version conflict inside \texttt{botocore}. It
pinned both packages, met a shared-object failure in \texttt{torch\_scatter},
and reached an AUPRC of 0.455 at step 67 against a published 0.474 $\pm$
0.025. 

That chain is not unusual. Two of the tasks make the agent build
third-party research code against a live environment, and at least one turn
failed in 72\% and 71\% of their runs. The remaining tasks read data we stage
in advance, and their rates fall to 27\% and 20\%. Recovery is usually
immediate. The turn after a failing turn succeeded 116 times out of 145. A
separate 9\% of runs ended early because the output of a dependency install
filled the context window. The extra spend buys that recovery, and the memory
keeps the command that produced it. We summarise the leaderboards, the split
files and the scoring procedure in the Methods section.

\textbf{Analysing biomedical data beyond any published protocol.}
We next moved from a value on record to the quality of an open analysis. Two
benchmarks carry that test between them. BiomniBench-DA\cite{Qu2026.05.12.724604} curates
data-analysis tasks that each ship a public dataset, a reference trace from a
domain expert and a rubric anchored at the decision points of that analysis.
Its authors score the process rather than the outcome because a correct final
answer can come from memorisation or from wrong reasoning that lands on the
right value. Fifty of its tasks have been released publicly, and those fifty
span 16 of its 17 task types. GeneBench-Pro\cite{Li2026.06.29.735386} scores
only the number at the end. Each of its problems gives a short context and a
target estimand, and a plausible wrong turn at any fork changes every step
that follows. Its authors report that models identify
local diagnostic signals and then fail to carry the implication through to the
matching analysis decision. No repository implements either set of tasks, and
the explore agent built one skill for each task type.

\begin{figure}[tp]
\centering
\includegraphics[width=0.95\linewidth]{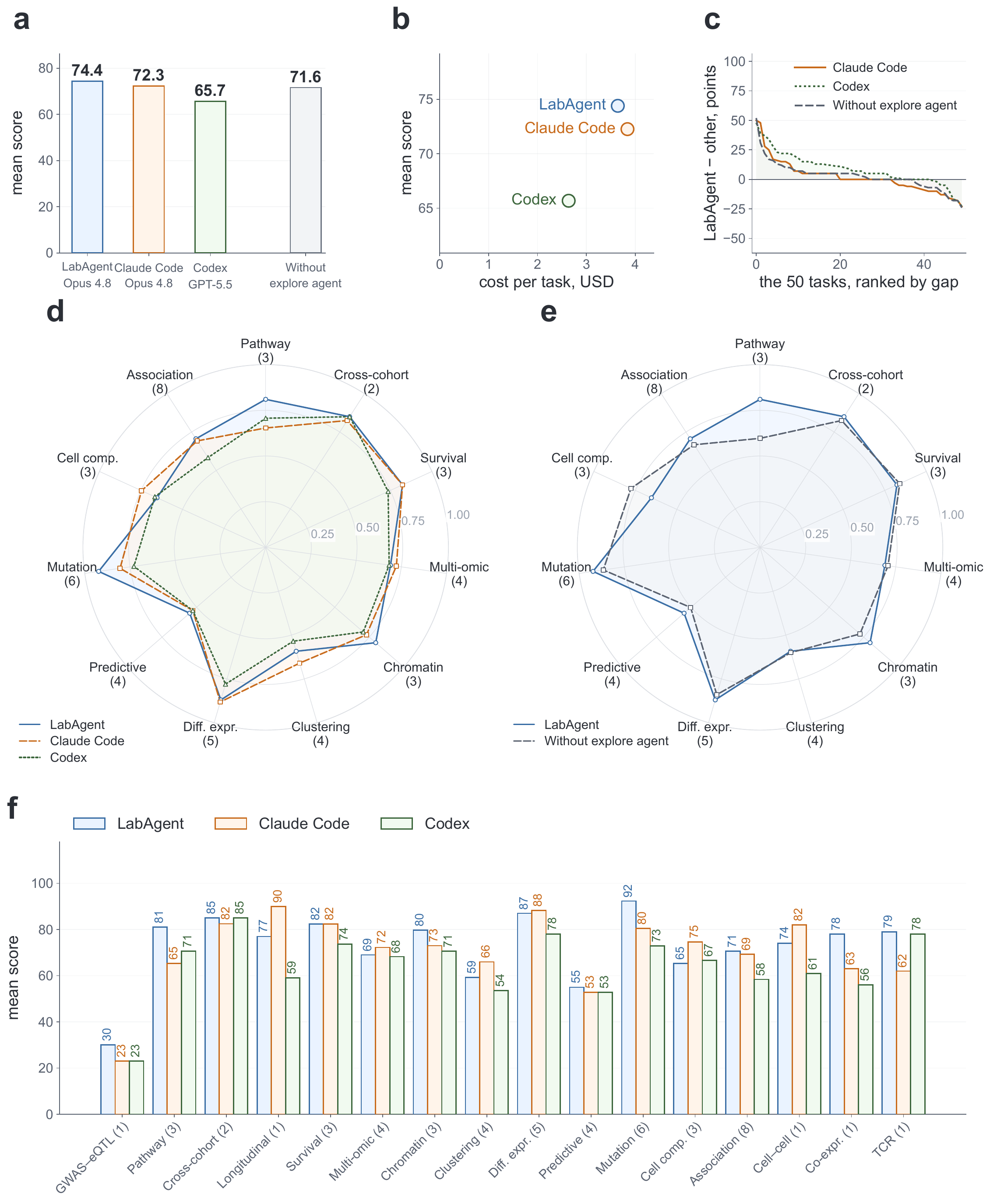}
\caption{\textbf{Open biomedical data analysis under a process-level rubric.}
(a) Mean rubric score across the 50 public tasks for \method{},
Claude Code, Codex and the arm with the explore agent removed.
(b) Mean score against cost per task.
(c) Per-task difference in score between \method{} and each other arm, with the 50 tasks ranked by the size of the gap. A value above zero marks a task where \method{} leads.
(d) Mean score by task type for the three systems, on a scale from 0 to 1, with the number of tasks beside each type.
(e) The same view for \method{} and the arm with the explore agent removed.
(f) The same scores as (d) on the rubric scale, with the value above each bar.}
\label{fig:biomni}
\end{figure}

First, we compared the three systems on the mean rubric score.
\method{} scored highest of the three at 74.4 (Fig.~\ref{fig:biomni}a). It
leads Claude Code by 2.1 points and Codex by 8.7. Codex also runs a different
base model, so the wider gap measures the harness and the model together and
attributes nothing to either alone. The two Opus arms trade places task by
task. \method{} led on 20 of the 50 tasks, trailed on 17 and drew level on 13
(Fig.~\ref{fig:biomni}c). Those per-task gaps run past 10 points in both
directions, so the 2.1 points at the mean hide a much wider spread. The
breakdown by task type puts our gains on mutation analysis and pathway
enrichment and puts Claude Code ahead on clustering and cell composition
(Fig.~\ref{fig:biomni}d,f). Finally, we ablated the explore agent and left a
reproduce agent with no skill library on this benchmark. That arm scored 71.6
and still placed above Codex (Fig.~\ref{fig:biomni}a). Against that arm, 
the full system scored 2.8 points higher with a standard error of 1.7 (Fig.~\ref{fig:biomni}c,e).

\begin{figure}[tp]
\centering
\includegraphics[width=0.9\linewidth]{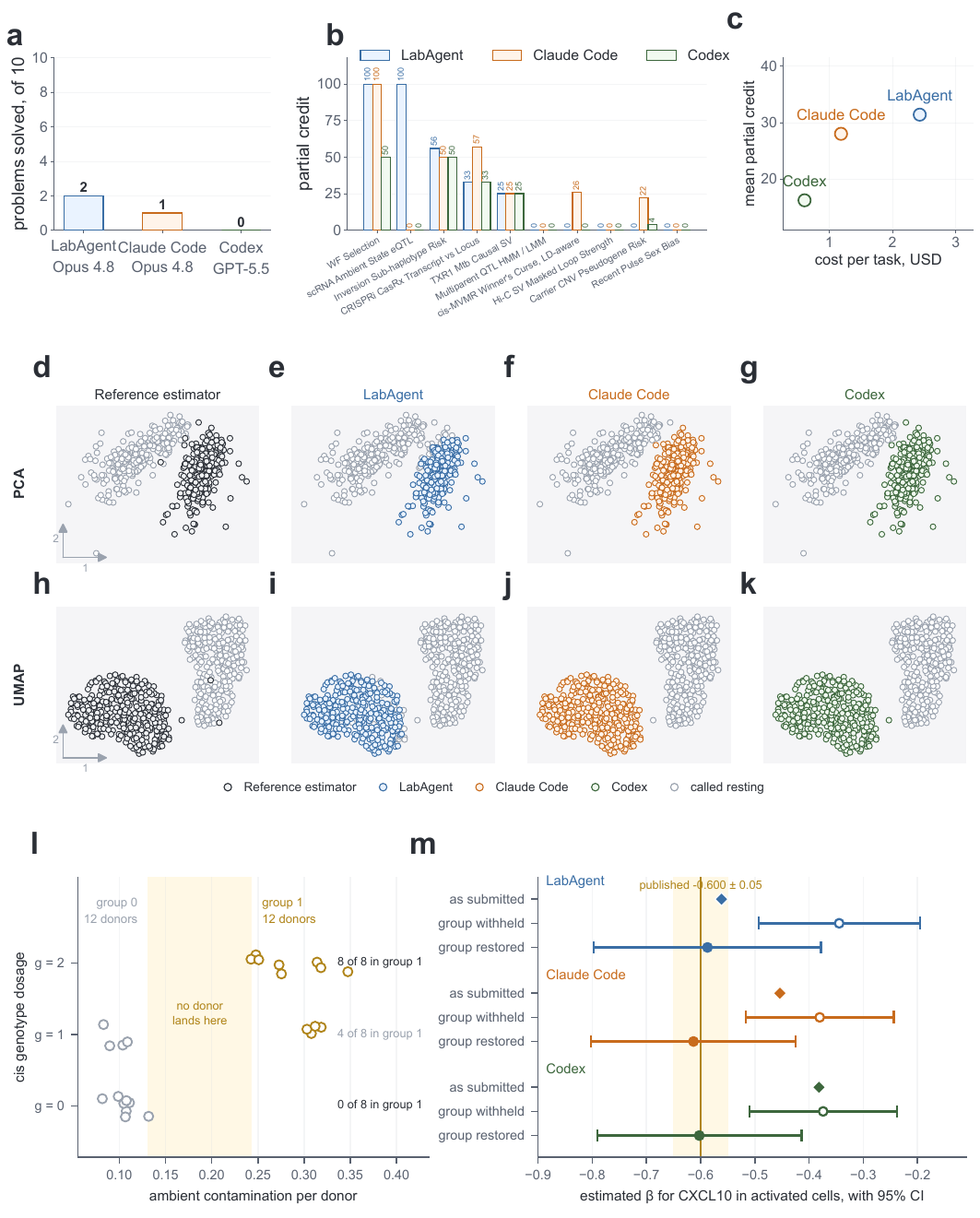}
\caption{\textbf{Open analysis across dependent decision points.}
(a) Problems solved outright, of 10.
(b) Partial credit on each of the 10 problems.
(c) Mean partial credit against cost per problem.
(d)-(m) The ambient-state eQTL problem.
(d)-(g) The 588 cells in the first two principal components, for
the reference estimator and then for each system. Cells the panel calls
activated carry its colour and the rest stay grey.
(h)-(k) The same cells in a UMAP embedding, in the same order.
(l) Ambient contamination for each of the 24 donors against its cis
genotype dosage. The shaded band holds no donor and separates the two
technical groups.
(m) The estimated effect on CXCL10 in activated cells with its 95\%
confidence interval, for each system as submitted, with the technical group
withheld from that system's own pipeline, and with the group restored. The
vertical line and the band around it give the published value and the
tolerance the grader allows.}
\label{fig:genebench}
\end{figure}

On GeneBench-Pro, \method{} passed 2 of the 10 public problems outright,
Claude Code passed 1 and Codex passed none (Fig.~\ref{fig:genebench}a).
Partial credit averaged 31.4 for \method{} against 28.0 for Claude Code and
16.2 for Codex, at a mean US\$2.43 per problem (Fig.~\ref{fig:genebench}b,c). The Codex figure
is an equivalent value under a subscription and not a metered bill. This
setting stayed the hardest of the five we ran, and five of the ten problems
returned zero for \method{}. One problem separates the three systems. The task asks for the cis expression quantitative trait locus (cis-eQTL) effect on CXCL10 in activated monocytes, and it supplies
single-cell counts that ambient RNA has contaminated. CXCL10 is an
interferon-inducible chemokine, and monocytes produce it in quantity once they
activate. The task therefore asks for a regulatory effect for one cell
state, and a measurement that pools the states averages that effect away. The
published estimate is $-0.600$ and the grader allows 0.05. \method{} returned $-0.561$, at 0.77
of that tolerance. Claude Code returned $-0.454$ and Codex returned $-0.382$,
at 2.92 and 4.36 of tolerance.

All three systems recovered the activated population from the same 588 cells
and landed close to the reference call (Fig.~\ref{fig:genebench}d--k).
Agreement reaches 0.951 for \method{}, 0.995 for Claude Code and 0.997 for
Codex. A hand-built agent therefore separates monocyte activation states as
well as either commercial harness does. \method{} agreed least of the three
and missed 29 of the activated cells. That gap does not propagate. We swapped
each system's state call into the reference pipeline and left every other step
alone, and all three landed inside the grader's tolerance. Any of the three
state calls supports the published answer, so the difference between the
submitted estimates arises at a later step. A donor-level variable the task
never mentions accounts for it. We measured ambient contamination for each
donor and grouped the 24 by what we found. They fall into two groups, and no
donor sits between 0.13 and 0.24 (Fig.~\ref{fig:genebench}l). Those groups
follow the cis genotype. None
of the 8 donors at dosage 0 sits in the contaminated group, 4 of the 8 at
dosage 1 do, and all 8 at dosage 2 do. Ambient RNA carries CXCL10, so a
donor's contamination adds a genotype-dependent amount to every cell we
measure. The contaminating term lines up with the genotype itself, and a
pipeline that ignores it returns a biased slope and not a noisier one. We
tested that directly. We held every other step fixed and varied the group term
alone. Each system falls outside tolerance once we withhold the group from its
own pipeline, and each returns inside once we restore it
(Fig.~\ref{fig:genebench}m). Every pipeline was already correct in every other
respect, so recovering the group is what separates a pass from a failure.

The three systems differed in what they did about that group. Claude Code
computed contamination for each donor, found it correlated with genotype at
0.77 and named that correlation as the task's central difficulty. It built its
ambient profile over the five measured genes instead of the empty-droplet
total, and that deflated every donor's contamination 4.85-fold. All 24 donors
fell below its threshold and the two groups disappeared, so Claude Code
modelled contamination as a continuous covariate on the mean. A linear term
fits a slope where the data show a step. Codex carried no group term anywhere.
\method{} ran two baselines of its own before it committed, one over all cells
at $+0.27$ and one over raw counts at $-0.011$. The estimate changed sign
between them, and \method{} read that reversal as evidence that its correction
had taken hold. We summarise the task selection for both benchmarks, the
skill-building mode and the two scoring procedures in the Methods section.

\textbf{Predicting protein variant effects without a named method.}
We then took a setting in which we withheld the method itself. Biologists predict the effect of a mutation to read what a protein
does, and dozens of published methods now compete at the task. A method leads
the ranking on one protein and falls past rank 70 on the next. A sequence-only
language model comes last of ten candidates on 48 of ProteinGym's 217 assays,
and an agent with no protein-specific reasoning reaches for that model before
any other.

\begin{figure}[tp]
\centering
\includegraphics[width=0.92\linewidth]{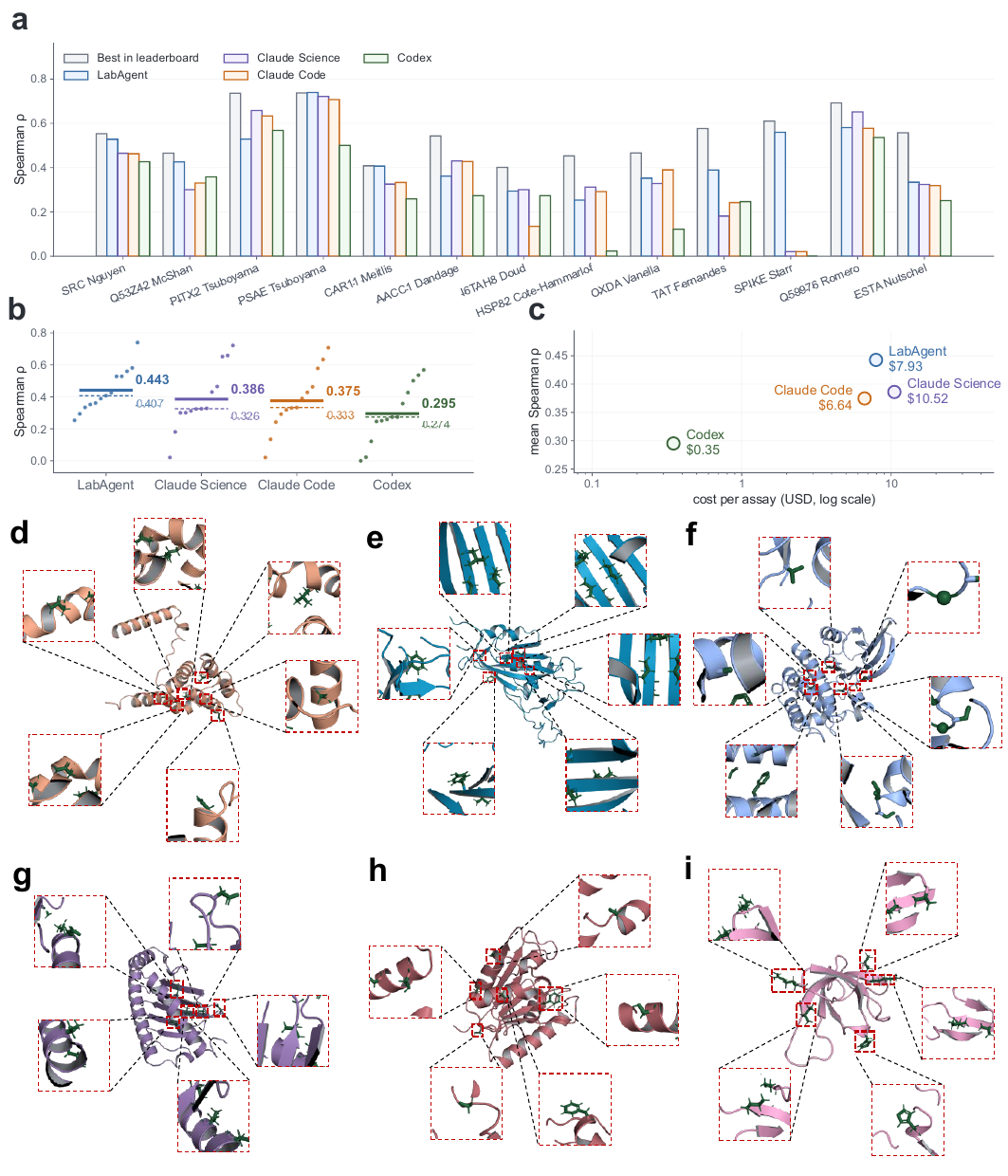}
\caption{\textbf{Variant effect prediction across 13 deep mutational scans.}
(a) Spearman correlation on each assay for the four arms. The dashed
line above each group gives the best published value for that assay.
(b) The 13 values for each arm, with the mean as a solid line and the
median as a dashed line.
(c) Mean Spearman correlation against cost per assay.
(d)-(i) Six of the scanned proteins. The cartoon carries the mean agreement between the reproduction and the measurement over each secondary-structure element, on one scale shared by the six panels that runs from $-0.35$ to $0.70$, and grey marks residues the scan did not measure. Sticks mark the positions of highest agreement that lie together in the fold, from four to six in each protein, and the inset enlarges that group. A glycine carries no side chain and appears in no panel, so panels (f) and (g) each draw one stick fewer than the text names.
(d) CARD11 CARD domain. (e) SARS-CoV-2 spike receptor-binding
domain. (f) SRC kinase domain. (g) HLA-A.
(h) \emph{Bacillus subtilis} lipase A. (i) Photosystem I
subunit PsaE.}
\label{fig:proteingym}
\end{figure}

We handed the agent the whole list in its published state. The explore agent
turned all 24 GitHub-hosted baselines in ProteinGym's acknowledgements table
into 26 skills. Some of the 26 skills are current, some have not been touched in
years, and some are famous and weak. Each arm received one assay, the whole
library and no method name. A filtered library would answer half of the
questions in advance and would leave every bad pick cheap. We fixed the assay
list by the same principle. Method choice decides the score on about
half of ProteinGym's 217 substitution assays\cite{notin2023proteingym},
and on the other half every reasonable
method lands in much the same place. A study that drew only
from the first half would manufacture its own result.
We stratified the benchmark by taxon and alignment depth, gave each stratum
seats in proportion to its size, and took the assay closest to the median
mutant count inside each one. That draw returned 10 assays and left one of
ProteinGym's own categories empty, so we ran the same rule once more over
those categories and added three. We removed nothing from the first draw,
since a sample discarded once its result is known is a sample chosen for its
result. The 13 assays sit on 13 different proteins. Neither pass looks at
which method wins, and both keep the assays where every arm ties. We scored
every reproduction at two levels. The first is the Spearman correlation across
a whole assay, and it averages over every position in the protein. The second
resolves each reproduction to single residues. Each scanned position carries
about 19 substitutions, and the agreement between the reproduction and the
measurement at that position follows from those 19 alone. We ranked every
well-covered position by that agreement. In each of six proteins we drew the
highest-ranked positions that lie together in the fold, and a group holds from
four to six of them. Every position we drew ranks inside the top 15 of its own
protein (Fig.~\ref{fig:proteingym}d--i).

\method{} reached a mean Spearman correlation of 0.443 and led all three
comparison arms (Fig.~\ref{fig:proteingym}a,b). It won 7 of the 13 assays
outright and never fell below 0.253 on any assay, while each of the other
three arms dropped below 0.10 on at least one (Fig.~\ref{fig:proteingym}b).
We traced the lead to a single decision. \method{} settled on one structure-aware 
predictor for 11 of the 13 assays and committed to it for almost the whole benchmark. 
Sequence-only models sit near zero on viral proteins, and \method{} led two of the three
viral assays by a wide margin. It reached the lead at US\$7.93 per assay
against US\$10.52 for Claude Science (Fig.~\ref{fig:proteingym}c).

The six panels fall into two groups by where the marked positions sit. In
three of them the positions are in the core of the fold. The SARS-CoV-2 spike
assay scored how much protein reaches the yeast surface, and that readout
reports on folding and not on receptor binding\cite{starr2020deep}. A central
beta sheet flanked by helices forms the core scaffold of that domain, the
receptor-binding motif sits on that scaffold, and four disulphides hold it
together\cite{starr2020deep}. C379 to C432 is the most important of the four
for the core\cite{starr2020deep}. All six positions we drew are hydrophobic or
aromatic, and five of the six sit below a relative solvent accessibility of
0.15 (Fig.~\ref{fig:proteingym}e). They fall into F392 to Y396 and V510 to L513 with I434
on its own. L513 sits 5.0 angstroms from C432, so the set reaches that most
important disulphide. By contrast, the receptor-binding motif holds 69 of the
201 scanned positions, and not one of the six falls inside it. An assay that
selects on folding marks the scaffold that folding rests on.

The other two of that group carry the same reading at the scale of a domain.
CARD11 is an adaptor protein that carries an antigen-receptor signal onward,
and the assay scored the variants that raise that signal without a
receptor\cite{meitlis2020multiplexed}. All six positions lie inside the
CARD, and four of the six sit below a relative solvent accessibility of 0.15
(Fig.~\ref{fig:proteingym}d). The gain-of-function variants reported in this
gene fall in the CARD, the LATCH and the coiled coil, and none of them falls
in the C-terminal domains\cite{zhao2022novel}. The SRC positions are the
most buried of the six, at a median relative solvent accessibility of 0.001,
and V405 lies two residues before the DFG motif
(Fig.~\ref{fig:proteingym}f). In the other group the positions are on the
surface. The \emph{Bacillus subtilis} lipase A positions are the most
exposed of the six, at a median relative solvent accessibility of 0.536, and
they lie 14 to 25 angstroms from the catalytic triad\cite{van2001crystal}
(Fig.~\ref{fig:proteingym}h). The photosystem I PsaE positions sit on the
surface of a 68-residue fold, at a median relative solvent accessibility of
0.481 (Fig.~\ref{fig:proteingym}i). The HLA-A positions sit between the two
groups at 0.245, they form one continuous surface, and none of them faces
the peptide-binding groove (Fig.~\ref{fig:proteingym}g). We read each panel 
against the features its own protein is known for, and we drew no rule across the six.

None of this survives a reproduction that lands on the wrong method. On the
spike assay Claude Code deployed a sequence-only ESM variant and reached
0.021. Its working directory holds the environment, the scoring script, the
log probabilities and the download log, and it executed what it chose without
error. Even so, the same code base ships an inverse-folding model at 0.465 and a
structure-aware model at 0.521, and the repository it selected already held a
better answer. Claude Science returned the same 0.021 here, and Codex exited with an
error on all three attempts. \method{} reached 0.560, within 0.050 of what its
own method reports, while the gap to Claude Code was 0.539.
Almost all of that gap came from the choice of method and not from how well
either arm ran it. Once no method is named the choice becomes the experiment,
and a wrong choice costs an order of magnitude more than any execution error
we measured. An assay measures one property of one protein, and no single predictor reads
every property well. \method{} holds a library of published predictors and picks
from it when the task names none. It can therefore help protein variant effect
prediction treat the choice of predictor as part of the experiment.
We give the library and the routing in the Methods section.

\textbf{Rebuilding a published figure in statistical genetics.}
We last asked for a whole published figure instead of a single number. A
figure is where a life-science paper puts the comparison a reader acts on,
and a group that has to pick a fine-mapping tool for its own cohort reads
the panels before it reads the text. Behind each panel sits a chain of
simulation, scoring, aggregation and drawing. Journals archive the image at
the end of that chain instead of the chain itself, and what survives is an
output and not the means to remake it. Data and code are the two inputs a
later group needs most, and the record falls short on both. One survey of
ecology found data for 79\% of articles and code for
27\%\cite{culina2020low}. The code that does survive often fails to run. One
study collected 27,271 Jupyter notebooks from 3,467 biomedical publications,
and of the notebooks it could rerun, 1,203 finished without an error and 879
of those returned what the notebook itself had
recorded\cite{Samuel2022ComputationalRO}. The SuSiE2
study\cite{zhang2024integration} sits on the better side of both counts. It
releases its simulation script in a public repository, and that script still
ends at a table of results. The article states the heritability grid, the
number of risk genes and the causal variants each one carries. It leaves the
prior variance, the correlation threshold, the number of single effects and
two properties of the eQTL layer to the script alone, and it leaves every
choice that turns the table into the figure nowhere at all. A later reader
inherits the simulation and not the figure.

\begin{figure}[p]
\centering
\includegraphics[width=0.95\linewidth]{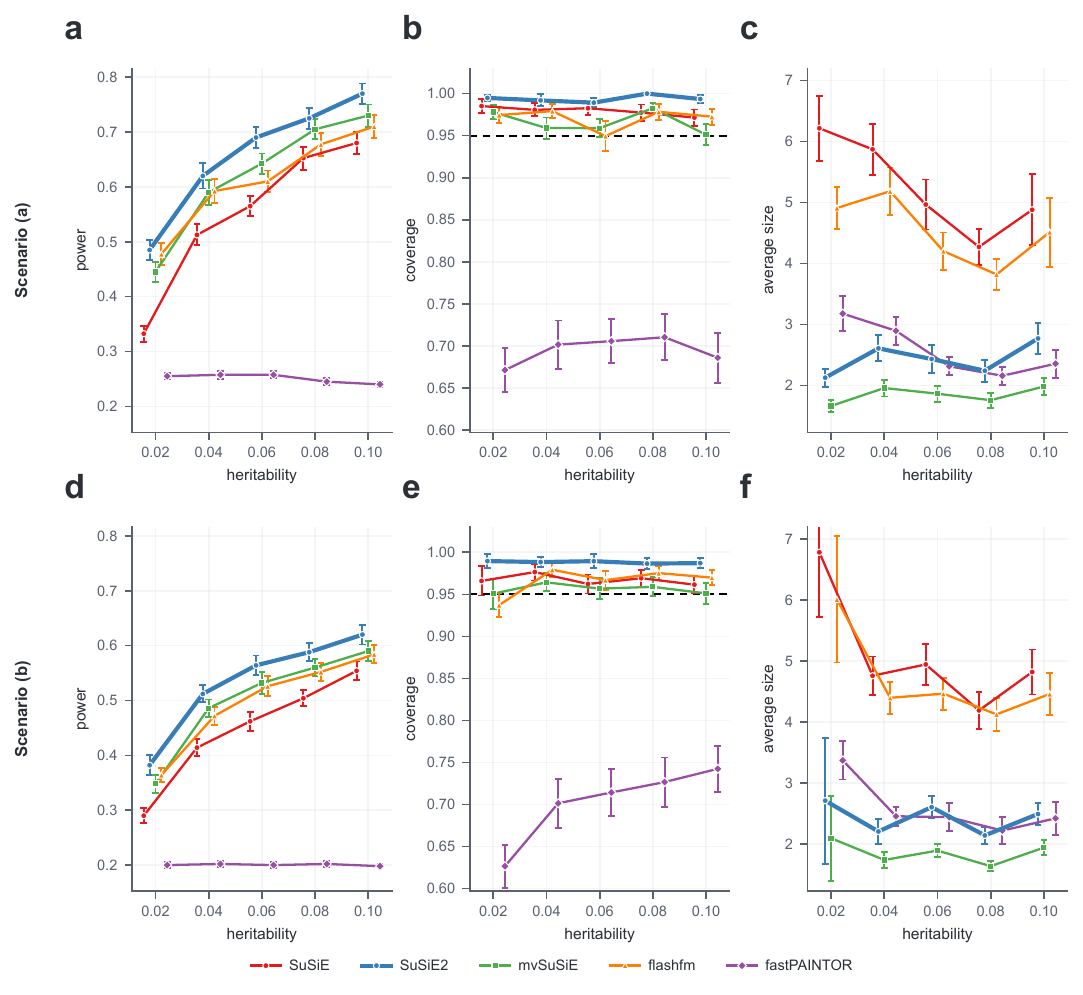}
\caption{\textbf{Rebuilding a published simulation figure.}
Five fine-mapping methods across five heritability settings, in two causal
architectures.
(a)-(c) The architecture in which every causal variant is also
an eQTL of a risk gene, at eight causal variants per locus.
(d)-(f) The architecture that adds two causal variants outside
the risk genes, at ten per locus.
(a),(d) Power, the share of causal variants that at least one
credible set captures.
(b),(e) Coverage, the share of credible sets that hold a causal
variant, with the nominal 0.95 as a dashed line.
(c),(f) Average credible set size.
Points are means over 50 replicates per cell against the 100 the article used,
error bars give the standard error across those replicates, and points sit
offset horizontally within each heritability so that overlapping intervals
stay visible. The SuSiE and SuSiE2 lines are the agent's own implementations
of the published algorithms, and the fastPAINTOR line is the supplied PAINTOR
binary. The mvSuSiE and flashfm lines are surrogates the agent wrote in place
of the published packages, with parameters it chose to match the reported
behaviour of those methods, and we show them for completeness and not as
reproductions. PAINTOR ran with a maximum of two enumerated causal variants.
That bounds its power at 0.250 and 0.200.}
\label{fig:susie}
\end{figure}

We followed the experimental configuration of that study and compared five
fine-mapping methods across two causal architectures and five heritability
settings. The five settings differ in what they read beside the association signal of
one trait. SuSiE reads nothing else\cite{Wang2018ASN}. PAINTOR adds
functional annotation\cite{kichaev2014integrating}, SuSiE2 adds
expression\cite{zhang2024integration}, and mvSuSiE\cite{zou2026fast} and
flashfm\cite{hernandez2021flashfm} add further traits. We scored the
relations the figure asserts alongside the quantities the article states.
The agent received the article, the authors' public repository, the PAINTOR
and FINEMAP\cite{benner2016finemap} binaries and a genotype set of 93,246
common variants on chromosome 1 for 10,000 samples. That genotype set came
from an earlier study by the same group, and an automated scan of the
finished run found no reference to the outputs that study had left on the
same cluster. The agent then built its five comparators in three ways. It
ported SuSiE from the published algorithm and checked the port on a separate
test locus, and it wrote SuSiE2 from the description in the text. It drove
the supplied PAINTOR binary directly. It wrote surrogates for mvSuSiE and
flashfm and never installed the published packages. Three lines therefore
carry the comparison. The repository holds no plotting code, and the agent
wrote its own to draw the six panels. Our task specification names the 100
replicates the article ran and permits fewer. The agent chose 50 per cell,
recorded that count beside every row of its output and finished the whole
grid in 2 h 40 min (Fig.~\ref{fig:susie}).

We read the three lines that carry the comparison against every ordering the
published figure asserts. Each one came back in the same direction. We found
SuSiE2 above every other method on power in all ten cells of the grid, and
it raised the detection rate over single-trait SuSiE by 11.1\% to 45.9\%
against the 15\% to 40\% the article reports. We then saw power rise with
heritability at every step for four of the five methods while the first
architecture stood above the second at every heritability for all five
(Fig.~\ref{fig:susie}a,d). We next read coverage against the nominal 0.95 at
0.962 to 1.000 for the two SuSiE lines but at only 0.627 to 0.742 for
fastPAINTOR (Fig.~\ref{fig:susie}b,e). We then asked what the grid says
about fine-mapping on real human data, and we took two readings from it. The
genotypes behind it are real. They come from 10,000 Europeans of the UK
Biobank, so the linkage structure the five methods work against is the
structure a real study meets. We noticed first that the expression layer
helps most where the association signal is weakest. At the lowest
heritability of the grid, SuSiE2 lifted power from 0.333 to 0.485, a gain of
45.9\%, and at the highest it lifted 0.680 to 0.770, a gain of 13.2\%.
Expression therefore matters most at the loci a single-trait analysis
handles worst. We noticed next how much no method reached. At the highest
heritability we ran, the best of the five recovered 0.770 of the causal
variants under the first architecture and 0.620 under the second. Real
linkage disequilibrium in a European cohort therefore still hides 23\% to
38\% of the causal variants from every method in the comparison, and the
harder of the two architectures is the one that packs more causal variants
into a single locus.

One departure traces to a single command-line argument. fastPAINTOR's power
stays flat across the grid, at 0.240 to 0.258 under the first architecture and
0.198 to 0.202 under the second, against gains of more than 0.20 for every
other method (Fig.~\ref{fig:susie}a,d). For that flat line the agent called
PAINTOR with its enumeration flag set to 2 and annotated that call in its own
source as the fast mode. Enumeration is the fast mode, and the 2 is a second
quantity, the largest number of causal variants the model may enumerate. Each
locus carries eight causal variants under the first architecture and ten under
the second, so the method could recover at most 0.250 and 0.200 of them. Mean
power came out at 0.2510 and 0.2004. The flag is documented, and this
departure is a misreading rather than a gap the article left open. Credible
set size also came back smaller than the article reports across the
comparison, and the run's README traces that to a purity and pruning rule the
agent set itself (Fig.~\ref{fig:susie}c,f).

Our rubric reads the output table against the claims the article states and
awards 88.1 of 100. That rubric sees none of the problems above. It cannot see
that two of the five lines are surrogates, and the agent chose the parameters
of those two to match conclusions the figure itself asserts, so about a third
of the total rests on lines that test nothing. The run's own record is not
exact either, since the README describes the fastPAINTOR credible sets as
built to 0.80 coverage while the call site leaves the default of 0.95 in
place. A reproduction that fails in ways a reader can name is what an audit
should deliver, and a summary score on its own would have shown none of this.
A credible set is the list of variants a later experiment has to test.
\method{} rebuilds the grid behind a published fine-mapping figure and states
the parameters that decide how long that list comes out. A later group can
therefore place a new method on the same axes as the published comparison and
take that comparison to harder fine-mapping settings. We give the scoring
rubric and the task specification in the Methods section. Overall, \method{}
contributes to reproduction and to discovery across drug property prediction,
genomic and single-cell analysis, protein science and statistical genetics.

\section{Discussion}
Reproducing and reusing published computational methods is important for any
science that builds on earlier work, and it is especially important for
biology. In the past, few reproduction pipelines were both reliable and
general. A laboratory releases only part of its code, one machine differs from
another, and an article leaves the parameters that matter unstated. With the
help of large language model agents, an executable skill library and a memory
that persists across runs, we introduce \method{} as a harness with a general
reproduction workflow for drug property prediction, multi-omics analysis,
single-cell eQTL estimation, protein variant effect prediction and statistical
fine-mapping. \method{} builds its skill library through a systematic search
and validation of released laboratory code, and its reproduction workflow
evolves on the experience that earlier runs accumulate. We tested \method{} on
three benchmarks and two case studies, and it outscored three commercial
harnesses on every setting that carries a comparison.

Regarding the contributions of \method{} to reproduction and discovery, we
showed that a route from released laboratory code to a validated skill is
workable, but a setting that already names the method it needs requires a
larger evaluation set before the gain from that route separates. However, most
questions in biology are open ones or questions of analysis and
interpretation, and they name no single best method. We also extend the
harness to open analysis settings such as BiomniBench-DA and GeneBench-Pro.
Although \method{} took the highest score of the three systems on both, it
will still help to ask how an agent of this kind evolves on longer and more
complex problems of biological analysis, and to raise prediction and
biological understanding through that evolution. It will also be valuable to
measure what the memory contributes on its own. We also note that about one
third of the entries came back above the published number, and that this rate
does not depend on how large an evaluation set the entry uses. Meanwhile, as
expected, a setting that names the method leaves the three systems hard to
tell apart, and a setting that withholds it opens the gap. Our experiments
cover both ends. The three systems reproduce the ADMET leaderboards to much
the same accuracy. On ProteinGym the choice of method costs far more than any
execution error does. We record the diagnosis of each failure together with
the command that resolved it, and we hand a later run a fix it can execute.
Each reproduction leaves the next one less to rebuild. Therefore, \method{}
can help us reproduce published computational methods and explore new
biological questions.

Previous reproduction efforts score the final number and thus leave the steps
in between unrecorded. Here, we kept the whole trajectory. It records what the
agent met at each step, what it diagnosed the problem to be, and which command
resolved it. The record carries the fix itself, and thus, it is a better
resource than an isolated number. Based on our analysis, failure concentrates
where third-party research code has to build against a live environment, and
recovery is usually immediate. Moreover, the experience we accumulate still
describes environments and tooling. We lack a comparable record for the
judgements an analysis makes, and thus, it will be helpful to design a
reusable form for those judgements.

Inspired by our observation that one reported number hides the steps an
analysis took, we consider utilizing \method{} to resolve the structure behind
a protein variant effect at more than one scale and to expose the variables a
single-cell analysis most easily mistakes. Based on our experiments, reading a
reproduction below its reported metric can reveal which part of a protein an
assay actually constrains. Therefore, \method{} can also be used to compare a
reproduction with the biology of the assay it reproduces, and not only with
the number that assay reported. Regarding the analysis of single-cell data,
\method{} recovered a donor-level variable that the task never names and that
decides whether the published estimate is reached. Therefore, \method{} can be
used to expose the variables an analysis of this kind most easily mistakes.

Despite the promising results of \method{}, there is still a big gap between
the reproduction of a number and an understanding of the biology behind that
number. Firstly, the technical variation in single-cell data often tracks the
biology under study. A donor-level variable can arise from the preparation of
a sample, and it can also carry real biological signal. We cannot tell in
advance which one will bias an estimate. It may be a feasible idea to build
control data with known confounders and to treat the recovery of such a
variable as a task that carries its own score. Secondly, the same gap reaches
published code. A laboratory usually releases its core model, and the path
from that model to the figure in the article rests on one-off scripts,
unstated parameters and manual preparation of data. We have no way to recover
that path. \method{} can infer a candidate path from the figure legend and the
methods section, and it can then check that path against the numbers a reader
takes off the figure itself. Moreover, when a task names no method, \method{}
tends to choose a method that is strong overall, and it skips the step that
matches a method to the biological question at hand. The strength of a method
on a general leaderboard is measurable. Its fit to one class of assay has no
established measure, and we lack a signal that rewards this step. The
selection pressure an assay applies and the evolutionary depth of a protein
can enter the selection step as explicit inputs. A separate evaluation of the
choice itself would give this step a target of its own. Finally, the relation
between the quantity an assay measures and the part of a protein a model
actually grasps remains unsettled. A current evaluation returns one
correlation coefficient for a whole assay, and it draws no distinction between
residues. We can not read off where a method holds its strength. A score that
separates positions by structure may open this question. It would weigh the
burial of each site, the domain it belongs to and its place on an interface
against the selection pressure of the assay.

Overall, \method{} is a useful tool for reusing published computational
methods from the released code of a laboratory and asking biological questions
that lie beyond any published protocol.

\section{Methods}
\textbf{Problem Statement.} Here we build \method{}: a framework of two
agents and a memory that persists across reproductions (Fig.~1). It takes a
laboratory's public code repository together with the study that repository
implements. It returns a validated executable skill and a reproduction of a
result the study reports. That reproduction sometimes exceeds the value the
study originally reported. We start from a repository URL and collect the
source code, the README and documentation, the tutorials and notebooks, the
configuration and environment files, and the paper the repository cites. We
write this collected set as $\mathcal{A}$, and from it an explore agent $E$
synthesises a skill $k$, a unit of procedural knowledge. A skill enters the
skill library $\mathcal{K}$ only if it passes an execution-based validation
$V$,
\begin{equation}
k = E(\mathcal{A}), \qquad
\mathcal{K} = \{\, k : V(k) = 1 \,\},
\end{equation}
so that every entry of $\mathcal{K}$ is executable rather than merely
readable. We write a reproduction target as
$\tau=(\text{paper},\text{repository},\mathcal{D},\mu,B)$. Here $\mathcal{D}$
is the staged dataset, $\mu$ the metric the study reports, and $B$ the budget
for the run. We give the components of $B$ and its per-task ceilings under
Task configurations. A reproduce agent $R$ then takes a skill $k \in
\mathcal{K}$ and executes $\tau$ within $B$. It works against a memory
$\mathcal{M}$ that persists across reproductions. It returns the value
$\hat{\mu}$ that \method{} itself obtains for the metric $\mu$, and an
updated memory,
\begin{equation}
(\hat{\mu},\, \mathcal{M}') = R\bigl(k,\, \tau,\, \mathcal{M};\, B\bigr).
\end{equation}
We judge the run against a reference value $\mu^{\star}$ and allow it a
deviation $\delta$ from that value. Those two form a scoring pair, and we
apply it only once the run has finished. A published study supplies the pair
as the value it reports and the uncertainty it publishes alongside that value.
A task that supplies its own answer key sets the pair by its own grading
procedure instead. The reproduction counts as verified when
\begin{equation}
\operatorname{ver}(\tau) \iff \bigl|\hat{\mu}-\mu^{\star}\bigr| \le \delta .
\end{equation}
In the first case the source therefore sets the threshold. A task that defines
repeated splits gives $\hat{\mu}$ as the mean over them. We require the agent
to report the per-split values and their dispersion alongside it, so that the
reproduction carries an error bar of the same kind as the published value.
Neither $\mu^{\star}$ nor $\delta$ is therefore an argument of $R$. Each
benchmark scores the run with its own procedure once the agent has terminated.
That procedure reads only the files the agent wrote and returns nothing to it.

\subsection{The \emph{Explore Agent}}
Our aim here is to turn $\mathcal{A}$ into a skill that is both admissible and
usable. We do so in four steps. We first ground the repository. We then draft
and reflect. Next we validate in tiers. Finally we split what passes into a
form that can be retrieved a piece at a time. Before any of this, however, the
agent has to be given something to work from. It accepts two forms of input.
Which one applies is decided by the target rather than by us. Some targets
name the implementation they are asking for. An ADMET leaderboard entry is one
such target. For these the laboratory's own repository is the right place to
begin. We hand that repository to the agent directly, and one repository
yields one skill. An open analysis task, by contrast, names no implementation.
Choosing a repository for it would be our judgement rather than the task's
requirement. In this second mode we supply only the task description and let
the agent find the candidates itself. Here it rewrites the task from several
expert perspectives, queries literature and code backends with each rewriting,
and then deduplicates and ranks what comes back. We issue the rewritings in
place of the task itself because a single query returns only a narrow view of
a literature. Meanwhile we bound this search and guard it against re-asking
the same question. We cap the rounds a supervisor may spend, we require a
written reflection before each one, and a supervisor may declare the search
finished only once every stop criterion in its own brief is met.
Which mode each task uses is stated under Task configurations. Either way,
what the agent holds at the end of this stage is a repository.

We first ground the skill in the repository rather than in the paper because
we want to capture how a method was actually run. The need for this is
practical, because a language model that writes integration code from the
paper alone produces identifiers and dependencies that do not exist. A survey
of 576,000 generations from 16 models found that 19.7\% of recommended
software packages do not exist\cite{spracklen2025we}. The same class of
fabrication has also been characterised at the level of functions, classes and
import paths\cite{Liu2024BeyondFC}. We therefore never draft a skill from
prose. We clone the repository first and extract an application programming
interface (API) snapshot from it. That snapshot parses what the code exposes.
It records the importable modules, the public functions and classes, the model
constructors and the usage examples that surround them. It also records the
declared dependencies and the repository metadata. We keep that snapshot
alongside the skill as an implementation contract, and constrain the drafting
model to use only identifiers that appear in it. This excludes fabricated
identifiers at the point of generation. Downstream validation never has to
discover them later.

We have now fixed what a skill may refer to but not what it may claim. A
skill, after all, does more than name the calls a repository exposes. It has
to tell the agent what to do at every quantitative step. Here a draft can
transcribe the repository faithfully and still be wrong. We need the agent to
arrive at the right number when it reads a skill, and authoritative prose is
not the same thing. These two come apart in one specific way. An assertion the
agent cannot test overrides the agent's own reading of the data with the skill
author's. The agent has no way to notice when that assertion is wrong. We
accordingly treat untestable assertions as worse than silence, and we pass
every draft through a second model call that a rubric governs. That rubric
demands a \emph{check} at every quantitative step the skill directs. A check
computes a quantity, prints it, and chooses between two courses according to
what it shows. An instruction, by contrast, names that course outright and
leaves nothing to check. We cover eight situations in which that distinction
decides the answer. The first four require the skill to check direction before
it derives a grouping or a composite score, to derive any threshold from the
observed distribution, to report how far it extrapolates, and to declare what
assumption several estimators share. The remaining four require it to parse
answer-key semantics literally, to assert the algebraic relations between
answer keys, to verify grouping indices and join counts, and to audit units
and scale. The reflection pass injects those that apply. Meanwhile we also constrain
what
reflection is allowed to add. No specific method, estimator, threshold,
package or parameter setting may be named as the correct one, unless the
grounded materials contain checkable evidence for it. We can trace one case
end to end to show that this works. One rubric item states an abstract
requirement only, and names no gene, task or method. The generated skill turns
it into a concrete check that groups provisionally on the most confident
marker. It then prints every other marker's ratio between the two groups, and
flags those that run the opposite way. One marker in that task is in fact
depressed more than tenfold rather than elevated, and that reversal had
produced the wrong coefficient in earlier runs. Finally, the same pass removed
a rule the model had invented for itself in earlier drafts. We had measured
that rule as net-negative against deterministic graders.

We then require execution rather than assertion before a skill enters
$\mathcal{K}$. A skill can read well and still not run. Most code that a paper
releases will not run as it arrives. One survey took 15,817 Python notebooks
from biomedical publications and tried to execute them. Only 7.61\% ran
without error, and 5.56\% reproduced the results those papers had reported.
The remainder raised exceptions, and these were attributable above all to
incompletely recorded dependencies\cite{Samuel2022ComputationalRO}. We accordingly put
every draft through three escalating tiers before we admit it. Tier~0 checks
the skill file itself. It asks whether the package exists in the software
index, whether the code repository resolves, and whether the package identity
matches the domain the skill claims. It then checks the syntax of every code
block and the consistency of its imports. Tier~1 then resolves dependencies
with a package-manager dry run, and confirms that each call in the quick-start
section exists in the installed module. Tier~2 finally creates a temporary
environment, installs the dependencies, and executes the quick-start end to
end. The agent rewrites only the offending section when a tier fails. It works
from the snapshot and the observed error, and then revalidates. Only skills
that clear all three tiers enter $\mathcal{K}$.

We now have a skill we can trust but not yet one the agent can use in one
piece. It carries a check for every quantitative step it directs and therefore
describes the procedure end to end. The agent, however, is only ever at one
step of that procedure. The whole document therefore puts most of what the
agent reads out of reach of what it is doing. The part that does apply must
also compete for context with the data. We cut the skill instead along the
same joints the procedure has. We write each skill as a root document of
roughly 50 lines. That document carries the task statement, the answer schema,
the hard rules that hold throughout, and an index. We hold the substance in
four to seven \emph{loci}. Each locus covers one step of the procedure, and
the agent retrieves it by name when it reaches that step. We also ship a
machine-readable index alongside the directory. We stage that whole directory
into the working directory before a run, and we refuse any run whose staged
loci come up empty. A skill without that check degrades silently to its 50
lines of pointers while the run still completes normally.

\subsection{The \emph{Reproduce Agent}}
For reproduction, the obstacle is less often the science than the machine. In
our own runs, described below, at least one command fails in roughly seven
runs out of ten wherever a method's code has to be built against a live
environment. Where the data is already staged, fewer than three runs in ten
fail at all. The reproduce agent therefore has to make a chosen method yield a
number on a machine that is not the one the skill was written on, and to do it
inside a finite budget. We break this into four steps. The agent first selects
a skill and then builds an environment for the machine in front of it. It next
works the skill's procedure through whatever failures arise. It finally
submits what it obtains to a comparison made outside the loop.

We take selection first because it is a harder decision than it looks. A
library may hold dozens of skills, while a single target should use exactly
one of them. A wrong choice also looks exactly like a poor execution from the
outcome alone, since both end in no usable number. We make selection a
separate and deterministic decision of its own for this reason. We show the
agent a catalogue of the descriptions of every skill in $\mathcal{K}$, and we
then ask it to return one skill name in a single call at temperature zero. We
match that reply against the library exactly. A name that matches nothing
raises an error, and no default catches it. A failed selection therefore
surfaces, and the system never absorbs it silently. We fix the call in this
way, so the choice also stays repeatable. It does not become a second source
of run-to-run variance on top of execution.

We select a skill that somebody else wrote on a machine that is not ours. Its
instructions carry that machine's assumptions with them. We therefore
prescribe execution as an ordered sequence. That order exists to force the
machine to be read before anything is installed on it. We build the loop on the default agent of
mini-swe-agent in the SWE-agent line\cite{yang2024swe}. The agent acts through the shell, and it issues one
command per turn. We supply a system prompt and an instance prompt
around this loop, and we write two pairs of them. The first pair serves the
tasks that make the agent build third-party research code. Those are TDC
ADMET, ProteinGym and the fine-mapping study. The second pair serves the two
open-analysis benchmarks. The account below follows the first pair. We embed the
staged skill in the system prompt and lay out the six roles the loop takes in
turn (Fig.~1). We then name the dataset and the output contract in the
instance prompt. The first two roles read the machine and prepare it. A probe
reads it, and an environment and data stage then prepares what the run needs
and checks the staged files against what the skill expects. The next three
produce the result. A code stage writes whatever the skill does not already
supply, a runner executes the workflow, and a verifier computes the metrics
the skill documents and writes them to a results file. The last role recovers
from failure. An analyst reads the error and proposes an alternative whenever
one of the others fails. The probe reads the compiler, the accelerator driver
and toolkit, the interpreter, the package manager and the modules the cluster
exposes. The versions it finds then select the install commands for the
following stage from an explicit table. This builds the numerical stack for
the machine actually in front of the agent. Meanwhile we give each run its own
virtual environment and prepend it to the path. Packages that one run installs
therefore cannot reach the next. Runs without that isolation adapted their
code to whichever version happened to be installed, and they did not pin the
version the skill specified. That drift moved one reported error from 0.256 to
0.273. We did not use containers, because the compute environment does not
provide them.

We rarely reach a reproduction on the first attempt even with an environment
in place. A run therefore takes largely the shape of its failures. Here we
describe what that looks like in practice, and we draw on the 140 runs of the
four tasks we report below. A run runs as a single sequence of shell commands
and their outputs, because the loop offers only one action. The model reasons,
issues a command, reads what comes back, and repeats. We handle failure inside
that sequence. We direct the agent in the system prompt to read the error and
try an alternative before it gives up. That alternative may be a different
install channel, a missing dependency or a reformatted input. We deliver
experience from earlier runs at the same point, by the mechanism we describe
below. A median of 16 to 19 turns was enough across the 140 runs. The task
where the agent chooses among skills runs longer, and its median reaches 38.
Failure, meanwhile, is routine wherever third-party research code has to be
built against a live environment. The two tasks that require it saw at least
one failing turn in 72\% and 71\% of runs. The figures are 27\% and 20\% on
the two tasks where we stage the data in advance. Recovery, where it happens,
is immediate. A turn that does not fail follows 116 of the 145 failing turns.
The diagnose-and-retry loop therefore closes inside the trajectory in four
cases out of five. A run can also go wrong in more than one direction. It can
circle without progress, spend without bound, hang on a single command, or
simply take too long. A cap on any one of these would leave the others open,
so we cap turns, monetary cost, the time of any single command, and the
wall-clock time of the whole run. Whichever binds first ends the run. None of
the 65 runs on the largest task reached the 100-turn cap, however, and one
reached the cost cap. The context window ended runs early instead, and runs
exceeded it in 9\% of cases because dependency installation produces very long
output. Finally, we do not take the agent's word for whether a run delivered a
result. Its self-report is unreliable in one direction, and no agent declared
failure in any of the 140 runs. We read the outcome from the harness's exit
status instead.

We do not treat a run that terminates without error as a reproduction by
itself. We therefore compare against $\mu^{\star}$ only after the agent has
stopped, and we leave that comparison to each benchmark's own scoring
procedure. That procedure reads only the files the agent wrote and returns
nothing to it. No part of it is reachable from inside the loop while the agent
runs.

\subsection{The \emph{Reproduction Memory}}
Nothing an agent works out inside one run reaches the next unless we write it
down, and without a store every run would begin from the same ignorance. Our
memory $\mathcal{M}$ is that store. We keep four kinds of record in it
(Fig.~1): the skills themselves, the grounding evidence each skill was built
from, the execution record of past runs, and the diagnoses that pair a failure
with what resolved it. We read the four at different moments. Skills are read
when one is selected, and grounding evidence when a skill is repaired. The
execution record we keep for audit rather than consult automatically. Only the
diagnoses are read back while a run is in progress, and they are what we
describe here.

We record the specific command that got the earlier run past the error because
a diagnosis is only useful if the next run can act on it. Suppose, then, that
we recorded only the advice behind that command. The next run would have to
turn the advice back into a command, and that is the step where it fails. We
therefore keep the command exactly as it was typed. We also keep a count of
how often it has since worked and how often it has not. A fix that succeeded
once may have succeeded only on that machine, and later retrieval should be
able to tell the two apart. We call that record a \emph{lesson}, and it
carries five fields. It opens with a trigger, either a literal string or a
regular expression. It then carries the text of what resolved it, the command
that did so, the runs in which it has held, and its success and failure
counts. Lessons divide into two sorts by how widely they transfer. Operational
lessons concern environments and tooling, and they transfer across every
domain by construction. A wheel that will not build is the same obstacle
whatever the science downstream, and so is a module that will not import.
Recipe lessons, by contrast, record the configuration that reproduced one
particular result, and their scope is correspondingly narrow.

We then deliver lessons to the agent on two channels. The first arrives before
the run begins. We take the highest-scoring operational lessons, and we add
any recipe lesson whose trigger names the current skill or task. We format
them into a block and prepend that block to the task description. The second
arrives during the run itself. We execute every shell command through an
environment wrapper, and that wrapper matches the command's output against
every known trigger. A match appends the corresponding lessons to the output
the agent will read on its next turn. We added this second channel because the
first one arrives at the wrong time. Context that we supply before a run must
anticipate which failures will occur. A match on command output, by contrast,
delivers the record at the moment and at the site of the failure, in the
channel the agent is already attending to. Meanwhile we cap what the wrapper
appends, so that it cannot crowd out the output it annotates. We also mark it
as a hint. The agent may therefore disregard it, and the skill takes
precedence where the two disagree.

We finally grow the store through the same wrapper that delivers from it. That
wrapper recognises a fixed set of failure signatures. We tag each of them by
kind and by language. These span environment faults, missing modules and
packages, absent commands, file and schema errors, shape mismatches,
accelerator faults, build failures and timeouts. We weight that set towards
dependency failures because that is where reproduction fails. The notebook
survey above found that unresolved imports and missing modules alone accounted
for 41.65\% of execution errors\cite{Samuel2022ComputationalRO}. We record the pair as a
candidate lesson whenever a command matches a signature and the next command
returns success. That lesson associates the signature with the resolution. We
then hold candidates for the duration of the run and append them to the
persistent store afterwards. We tag each with whether the wider run succeeded,
so that later retrieval can weight them accordingly.

\subsection{Benchmarks and datasets}
A computational biologist needs several abilities. Each one rests on the ones
before it, and we chose one task for each. The first ability is to run a
published method on unfamiliar hardware and land on the number its authors
report, and we test it on TDC ADMET\cite{Huang2021TherapeuticsDC}. The second is to carry an
analysis from a dataset to a conclusion when nobody has written the procedure
down, and we test it on BiomniBench-DA\cite{Qu2026.05.12.724604}. The third is to see the
artefact in a dataset that nobody named, and we test it on
GeneBench-Pro\cite{Li2026.06.29.735386}. The fourth is to choose among many
published methods the one that matches what an experiment measured, and we
test it on ProteinGym\cite{notin2023proteingym}. The fifth is to rebuild the
experiment behind a published figure, and we test it on a fine-mapping study
from our own laboratory\cite{zhang2024integration}. These five abilities run from execution to design, and together they ask
how far a laboratory's own methods carry once they leave that laboratory.
The five tasks also spread across four domains of life science. TDC ADMET
sits in drug discovery. BiomniBench-DA and GeneBench-Pro cover genomic and
single-cell analysis. ProteinGym sits in protein science, and the
fine-mapping study sits in statistical genetics. Each task carries its own ceilings on turns, on cost, on the
time of any one command and on the wall-clock time of a whole run. They run
from 40 to 150 turns and from 5 to 40 US dollars, and we raise one only after a
run has died against it.

\textbf{TDC ADMET.} The ADMET leaderboards of the Therapeutics Data
Commons\cite{Huang2021TherapeuticsDC} rank published methods on 22 datasets. Those datasets cover
absorption, distribution, metabolism, excretion and toxicity (ADMET). These
entries are known to be hard to rebuild. Ten distinct models occupy the top
three places across the 22 leaderboards. An independent assessment found that
only three of them could be reproduced\cite{Koleiev2026CriticalAO}. The rest failed
on environments that would not build, or on splits that leaked test molecules
into training. We take the top three entries of each leaderboard, which gives
66 targets. Sixty-five of them remain reachable, since one repository has been
deleted upstream. Those entries resolve to 11 distinct repositories. We build
one skill from each of them in repository mode, so the library holds 11
skills. We stage each entry with a training and validation table, a test
table, and five seed splits, and all of these come from the leaderboard's own
files. The task text then names the method to reproduce, and quotes the
entry's published value together with its standard deviation. That value comes
from the leaderboard's own public record of the entry. Finally, the agent
trains on each of the five seeds and evaluates on the corresponding test
split. It then reports the per-seed scores with their mean and standard
deviation.

\textbf{BiomniBench-DA.} BiomniBench-DA\cite{Qu2026.05.12.724604} curates 100 data-analysis
tasks across 17 task types from 21 published studies. Each task ships the
study's public dataset together with a reference analytical trace that a
domain expert wrote. It also ships a rubric of five to ten criteria. Each
criterion sits at a decision point of that analysis. Fifty of the 100 tasks
have been released publicly, and those fifty are the ones we use here. They
span 16 of the 17 task types. We stage the data ourselves, so the task is one
of analysis. We build skills here in task-type mode, one for each task type.
The output contract requires both a written trace of what was done and an
answer file, because the rubric grades the analysis and not only its
conclusion.

\textbf{GeneBench-Pro.} GeneBench-Pro\cite{Li2026.06.29.735386} poses
multi-stage analyses from genomics and quantitative biology. It gives the
agent a brief context and a target estimand, and the agent must then find its
own way to the correct workflow through several dependent decision points. The
benchmark comprises 129 problems, and its authors have released ten of them
publicly. Those ten are the ones we use here, and its authors hold the rest
out. We again build skills in task-type mode, one per problem, because no
repository is named for these tasks and the candidate methods have to be found
first. We stage the data in advance, so the agent never has to build the
method. The agent writes a single answer file in a fixed JSON schema, and a
deterministic grader then compares the numeric answer against a stated
tolerance.

\textbf{ProteinGym Variant Effect Prediction.}
ProteinGym\cite{notin2023proteingym} assembles deep mutational scanning
assays across a wide range of proteins. Each assay measures the effect of
every single amino-acid substitution in one protein, and the benchmark
publishes the score of every method it catalogues on every assay. We configure
it here so that the method is not given. Our sampling rule has four parts.
First, we draw 10 assays by stratified sampling over taxon and alignment
depth, and we allocate places to strata by largest remainder in proportion to
each stratum's share of the full set. Second, we take the assay whose mutation
count is closest to the median within each stratum, and we break ties by the
assay identifier. Third, we constrain the sample to one assay per protein and
to at most two per source study. Fourth, we apply the same within-stratum rule
once more over ProteinGym's own assay categories and add one assay for each of
expression, activity and stability. That fourth part exists because the first
draw stratified on taxon and depth alone. It returned no assay at all from the
expression category and it over-drew organismal fitness, since the cap of two
per study also capped stability, which one campaign dominates. We added to the
first draw and removed nothing from it, because a sample discarded once its
result is known is a sample chosen for its result. The three assays the fourth
part adds are the SARS-CoV-2 spike expression scan, the Romero 2015 scan of
Q59976 and the Nutschel 2020 scan of \emph{Bacillus subtilis} esterase A. We
fixed all four parts before any method was run, and none of them consults a
method's performance. The library itself holds 26
skills, one for each GitHub-hosted baseline in ProteinGym's acknowledgements
table. We deliberately do not filter it by rank, because a library in which
every method is strong would answer half the question before the agent is
asked it. We stage each assay with its variant list, its multiple sequence alignment and its predicted structure. We also report how exposed a marked
position is. We take the same predicted structure, strip the hydrogens, and compute the solvent-accessible surface area of every residue by the Shrake-Rupley method at a probe radius of 1.40 angstroms and 960 sphere points. We divide that area by the theoretical maximum for the residue type in the extended Gly-X-Gly state\cite{tien2013maximum}, which gives a relative solvent accessibility. A value below 0.15 marks a buried position by the usual convention. Alignment-based, sequence-based and
structure-based methods are therefore all available to the agent. Meanwhile we withhold the measured effects from every system alike. The output contract
then asks for one score per variant. The task text states the required
direction identically for every system we compare.

\textbf{UK Biobank Fine-mapping Simulation.} The UK Biobank supplies the
genotypes behind the first figure of a published fine-mapping
study\cite{zhang2024integration}. That study comes from our own laboratory, and we stage the
genotype panel it assembled for the agent. We take the first figure of that
study as our target here. It reports a simulation over 10,000 Europeans from
that cohort. Each region in the simulation comes at random from chromosome 1
and carries 5,000 single-nucleotide polymorphisms (SNPs). The figure compares
five fine-mapping methods across two causal architectures and five
heritability settings. The figure reports three quantities for each
combination, namely the power to detect a causal SNP, the coverage of the
credible sets, and their average size. However a reproduction of figureis the
harder request\cite{miao2025paper2agentreimaginingresearchpapers}. It reports a pattern rather than a value, so
it often gives no single number to compare against. We give the agent the
article, that genotype panel, the repository the laboratory released, and two
fine-mapping executables. An experiment produces a figure, and a method
paper's repository releases the method. Reproducing the figure therefore asks
for what a repository of that kind is not meant to carry. An experiment fixes
a grid of conditions and a number of replicates, and it then turns what comes
back into a figure. We take that experiment as the reproduction target here.
Five of its settings appear in the repository as bare numbers. Two of them fix
the geometry of each region. One sets the width of the window around each risk
gene, and the other sets the spacing between those windows. Two more fix how
heritability divides between causal and background SNPs, once for expression
quantitative trait locus heritability and once for the phenotype. The fifth
fixes how the two causal SNPs of each gene divide between their two roles. The
student who ran the simulation has since graduated, and the agent therefore
has to work out what each of the five settings controls. It then has to
rebuild the experiment around a method it already has. This task measures
whether it can recover what a laboratory knew but never wrote down.

\subsection{Evaluations}
For each task we set out the scoring procedure below, alongside what it can
and cannot see, because the scoring procedure decides what that task can
detect. Every one of them runs after the agent has terminated, and reads only
the files the agent wrote. We judge each task in its own metric and its own
units, and average no score across tasks, since an error and a correlation
share no scale and improve in opposite directions. We report cross-task
summaries as counts instead. Tasks also differ in what they tell the agent
about $\mu^{\star}$. A leaderboard entry publishes its value as part of its
own public record, and the task text then quotes that value identically for
every system we compare. Other tasks, by contrast, withhold the ground truth
from every system alike.

\textbf{TDC ADMET.} We do not choose a tolerance of our own for
this benchmark. We take the one the leaderboard already publishes, since every
entry publishes the spread of its own result. The source therefore supplies
the threshold, and we never have to defend a number of our own. We score each
entry in the metric its own leaderboard uses. We then count a reproduction
when that score falls inside the standard deviation the leaderboard publishes
beside the entry's value. That spread lies between 0.5\% and 2\% of the value
for most entries, which is considerably stricter than a conventional tolerance
band. That bar, however, is also unforgiving of small numerical differences.
We add a ratio-scale reading as well. A reproduction counts as close on that
reading when it falls within 5\% of the published value. We report both
readings for every entry we attempt.

\textbf{BiomniBench-DA.} We have no single right number to score
an open analysis against. This benchmark therefore scores what an answer
contains rather than how far it sits from a target. Its own protocol has a
language model read the agent's trace and apply the rubric. That model awards
each criterion full, half or no credit. We score by a second route instead,
because a language model does not score deterministically. The same trace need
not receive the same verdict twice, and no part of that verdict can be traced
back to the criterion it came from. We therefore match the trace and the
answer against the wording of each criterion's fully-correct level, and we
score the overlap between them. That procedure returns the same value for the
same answer every time, and every point it awards can also be attributed to a
criterion.

\textbf{GeneBench-Pro.} The agent works through a chain of
dependent decisions in these problems. Proximity to the answer therefore
carries no information at all. An agent that takes the wrong fork early
produces a number of the correct order of magnitude that is nonetheless wrong.
The benchmark grades accordingly, with a binary verdict against a recoverable
target under a tolerance it calibrates for that problem. A verdict alone,
however, does not say how far a run got before it went wrong. We accordingly
record a partial score over the quantities each problem asks for. We report
that score alongside the verdict.

\textbf{ProteinGym Variant Effect Prediction.} The assay files
ProteinGym distributes carry the measured effect beside each variant. Skills
that we write in the reproduce-a-metric idiom then end with a correlation
between a prediction and exactly that column. We would therefore make the task
trivially exploitable if we staged those files unchanged. An arm that emitted
the measured effect as its own prediction would score a perfect correlation,
and its transcript would look entirely compliant. We therefore remove the
measured effects from the staged data, and we hold them in a directory no arm
is given. We score predictions by the Spearman correlation between the score
an arm assigns to each variant and the effect the assay measured for it.
However, we do pay a cost when we withhold the effects. An arm can no longer
check its own sign convention against the data. We state the required
direction in the task text instead, and we use identical wording for every
arm. Finally, we read two further quantities from the leaderboard for context.
We take the best score any skill in the library can reach on that assay, and
we take the best score any method has published.

\textbf{UK Biobank Fine-mapping Simulation.} A figure leaves a
reader with a shape rather than a number. It says which method leads, which
way a curve moves, and which condition is the easier one. A shape does not by
itself measure anything, however, and a comparison of images would test the
plotting rather than the science. We therefore score the figure twice over,
once for the relations it asserts and once for the quantities the article
states beside it. We divide a hundred points among eight claims in three
groups. Forty points go to the relations the figure asserts. The figure says
that SuSiE2 leads on power, that power rises with heritability, and that the
first causal architecture is the easier of the two. Another forty go to the
quantities. We ask how large the gain over the alternatives is, how well the
credible sets cover, and how large those sets are on average, and we compare
each answer against the value the article reports. The last twenty ask whether
the work exists at all. We check that the grid of conditions is complete and
that the deliverables the article names are present. Weights within each group
follow how central a claim is to the published argument, and the largest single
weight, 30 of the hundred, goes to the claim that the method the article
introduces leads on power. We derive the rubric from the article itself, and we
apply it only once the run has finished. We own the target study, so its
precomputed outputs sit on the same cluster and can be read there. A separate
check therefore scans the run for any reference to them.

\subsection{Baselines}
We use Opus 4.8 as the base model of \method{}. We compare it against
three systems that each receive the same task text word for word, the same
staged data and the same budget ceiling. Claude Code CLI runs on Opus 4.8, the
same base model, whereas Codex CLI runs on GPT-5.5. Claude Science is a
workbench for scientific work. It ships more than sixty scientific databases
and toolkits, spanning genomics, single-cell analysis, proteomics, structural
biology and cheminformatics. Within it a coordinating agent delegates to
specialist sub-agents, and a separate reviewing agent then checks citations
and calculations. We run it on ProteinGym, where a system that already arrives
with a library of methods is the most informative comparison available. A
comparison means most when the systems differ in one respect and agree in
every other, and that is not what these comparisons are. Each of the three is
a complete system in its own right, not \method{} with a part removed.
\method{} and Claude Code share a base model and differ in everything
else, and \method{} and Codex differ in the base model as well. No
pairwise difference among them therefore attributes an outcome to any single
design choice, and what they give is a comparison of systems as they are built
and priced. Attribution instead needs a fourth arm, in which we run
\method{} against itself with the explore agent removed. That arm holds
the harness, the base model, the task text and the budget fixed, and varies
only whether a skill library exists. A difference between it and the full
system is therefore a difference the library made.

\subsection{Statistics and Reproducibility} 
We attempt each target once. Every value we report is therefore a point
estimate at the level of the run. A task may nonetheless call for repetition
inside that run. The ADMET tasks require five seed splits, so a value reported
for one of those targets is a mean over five fits and carries a dispersion of
its own. We compare two systems with a paired $t$-test over the tasks that
both completed. For every such comparison we give the number of pairs, the
mean difference, its standard deviation, the statistic and the $p$-value. We
report these even where the difference is not significant. We also fixed two
choices before any system ran. The stratified rule that picks the ProteinGym
assays looks at no method's performance, and the measured effects were removed
from the staged assay data for every system alike. We exclude one ADMET target
throughout, an entry whose repository has been deleted upstream and which no
system can attempt. Both stages of \method{} use the same base model. We
run skill selection at temperature zero so that the choice is repeatable, and
skill drafting at temperature 0.3, retrying up to three times when the model
returns malformed output. The runs themselves were submitted as scheduler
array jobs sharded round-robin across nodes, on the execution loop of
mini-swe-agent 2.4.1. Finally, the 65 reachable ADMET targets cost US\$296.13
in total, a median of US\$2.76 and a mean of US\$4.56 each.

\subsection{Data Availability}
The Therapeutics Data Commons ADMET benchmark group and ProteinGym are openly
available from their respective projects. So are the public releases of
BiomniBench-DA and GeneBench-Pro. We read the ADMET leaderboards on 2026-06-07, we
confirmed the ProteinGym baseline repositories on 2026-08-17, and we use
BiomniBench-DA and GeneBench-Pro as their projects released them. The genotype panel behind the fine-mapping simulation
was assembled from UK Biobank data for an earlier study from our own
laboratory\cite{zhang2024integration}, and we stage that panel unchanged. The access code for UK Biobank data is 29900. Individual-level
UK Biobank data cannot be redistributed, and are available to approved
researchers through the UK Biobank Access Management System. The generated
skill libraries and the reproduction trajectories that support the findings of
this study are provided as Supplementary Data.

\subsection{Code Availability}
We used the resources from the Yale High Performance Center (Yale HPC) to conduct all of the experiments. The codes of \method{} can be found in \url{https://github.com/fpxlei/LabAgent} under MIT license.

\section{Acknowledgments}

T.L. acknowledges the support of OpenAI Researcher Access Program.

\section{Author contribution}

L.L. designed this study with T.L. L.L. implemented the method and performed experiments. L.L., T.L., Y.Z. performed analyses. All authors reviewed and revised the manuscript. T.L. and H.Z. supervised this project.

\bibliographystyle{unsrt}
\bibliography{References}

\appendix
\clearpage
\setcounter{figure}{0}
\setcounter{table}{0}
\renewcommand{\thefigure}{\arabic{figure}}
\renewcommand{\thetable}{\arabic{table}}
\renewcommand{\figurename}{Supplementary Fig.}
\renewcommand{\tablename}{Supplementary Tab.}

\definecolor{siInk}{HTML}{2B2F36}
\definecolor{reproBar}{HTML}{C86B1C}
\definecolor{reproBody}{HTML}{FFF4EA}
\definecolor{explBar}{HTML}{3D673D}
\definecolor{explBody}{HTML}{F1FAF0}

\lstdefinestyle{prompt}{
  basicstyle=\ttfamily\scriptsize\color{siInk},
  breaklines=true,
  breakatwhitespace=false,
  postbreak=\mbox{$\hookrightarrow$\space},
  columns=fullflexible,
  keepspaces=true,
  showstringspaces=false,
  aboveskip=0pt,
  belowskip=0pt,
  literate={\ }{{\ }}1
}

\newcounter{promptlisting}
\tcbset{caseboxbase/.style={
  breakable, enhanced, listing only,
  listing options={style=prompt},
  coltitle=white, fonttitle=\bfseries\footnotesize,
  boxrule=0.7pt, arc=2pt,
  left=5pt, right=5pt, top=4pt, bottom=4pt,
  toptitle=1.5pt, bottomtitle=1.5pt,
  before skip=1.1em, after skip=1.1em,
}}
\newtcblisting{casebox}[3]{caseboxbase,
  colback=#1Body, colframe=#1Bar,
  title={\refstepcounter{promptlisting}%
         Supplementary Listing~\thepromptlisting.\hspace{0.5em}#2\label{#3}}}

\section*{Supplementary Information}

\lstdefinestyle{prompt}{
  basicstyle=\ttfamily\scriptsize,
  breaklines=true,
  breakatwhitespace=false,
  postbreak=\mbox{$\hookrightarrow$\space},
  columns=fullflexible,
  keepspaces=true,
  frame=single,
  framesep=4pt,
  rulecolor=\color{gray!50},
  showstringspaces=false,
  captionpos=b,
  aboveskip=1em,
  belowskip=1em,
  literate={\ }{{\ }}1
}

\subsection*{A. Prompts and the acceptance rubric}

The Methods describe what each prompt does. We show every prompt here
unchanged, with its Jinja tags intact, so that a reader can see which fields
the harness fills at dispatch and which text is fixed.

The reproduce agent works from two pairs. The first pair serves TDC ADMET and
ProteinGym, where the agent has to build third-party research code, and its
system prompt opens with a probe of the machine whose findings select the
install commands that follow. The second pair serves BiomniBench-DA,
GeneBench-Pro and the fine-mapping study, and it is built around skill loci the
agent reads on demand, an answer file in a fixed schema and a deterministic
checker. A caller can replace the answer contract through one argument, and the
fine-mapping run did so, since its scorer reads a table of metrics rather than a
JSON answer.

\begin{casebox}{repro}{Reproduce agent, first pair, system prompt.}{lst:repro_v1_system}
You are an autonomous experiment reproduction agent running on an HPC cluster (Linux, no Docker available, conda may be available).

Your task is to reproduce an experiment using the **{{ skill.name }}** tool by following its documentation exactly.

---

## Tool Documentation: {{ skill.name }}

{{ skill.description }}

{{ skill.full_markdown }}

---

## Operating Constraints

- You have access to ONE action type: `bash_command`. Issue one bash command per turn.
- Working directory: `{{ work_dir }}`
{% if conda_envs_root %}- Conda environments root: `{{ conda_envs_root }}`
{% endif %}- Do NOT use Docker or Singularity. Use conda/pip/source installs as documented.
- After each command, check the exit code and output before proceeding.
- If a step fails, read the error message and try an alternative (different install path, missing dependency, reformatted input) before giving up.
- Print informative progress messages so the trajectory is easy to follow.

## Workflow (follow in this order)

### Step 0 -- System Probe (ALWAYS run first)

Run the following diagnostics before touching any environment:

```bash
echo "=== GCC ===" && gcc --version 2>/dev/null | head -1 || echo "gcc not found"
echo "=== CUDA (nvcc) ===" && nvcc --version 2>/dev/null | grep "release" || echo "nvcc not found"
echo "=== CUDA (nvidia-smi) ===" && nvidia-smi 2>/dev/null | grep -E "CUDA Version|Driver Version" || echo "nvidia-smi not available"
echo "=== Python ===" && (python --version 2>/dev/null || python3 --version)
echo "=== Conda ===" && (conda --version 2>/dev/null || echo "conda not found")
echo "=== HPC Modules ===" && (module avail 2>&1 | grep -iE "pytorch|cuda" | head -10 || echo "module command not available")
```

Record the **CUDA version** (e.g., 11.7, 12.1) and **GCC version** (e.g., 8.5.0, 12.2.0) from the output.
These determine which PyTorch, torch_geometric, and other GPU-dependent packages to install.

---

### Step 1 -- Environment Setup

Create (or reuse) a conda environment, then install dependencies using version-appropriate commands
derived from the Step 0 probe.

#### 1a. PyTorch (version-aware)

Choose the install command that matches the detected CUDA version:

| Detected CUDA | Install command |
|---|---|
| 11.7 | `pip install torch==2.0.1+cu117 torchvision==0.15.2+cu117 --index-url https://download.pytorch.org/whl/cu117` |
| 11.8 | `pip install torch==2.1.2+cu118 torchvision==0.16.2+cu118 --index-url https://download.pytorch.org/whl/cu118` |
| 12.1 | `pip install torch==2.1.2+cu121 torchvision==0.16.2+cu121 --index-url https://download.pytorch.org/whl/cu121` |
| 12.2 - 12.4 | `pip install torch==2.3.0+cu121 torchvision==0.18.0+cu121 --index-url https://download.pytorch.org/whl/cu121` |
| No GPU / CPU | `pip install torch torchvision --index-url https://download.pytorch.org/whl/cpu` |

**HPC shortcut:** If `module avail` listed a matching PyTorch module (e.g. `PyTorch/2.1.2-foss-2022b-CUDA-12.1.1`),
load it with `module load <name>` instead of pip-installing to avoid a redundant download.

#### 1b. PyTorch Geometric (torch_geometric / pyg)

After PyTorch is confirmed installed, run:

```bash
# Detect the exact torch version string and CUDA tag that was installed
TORCH=$(python -c "import torch; print(torch.__version__)")
CUDA=$(python -c "import torch; v=torch.version.cuda; print('cu' + v.replace('.','') if v else 'cpu')" 2>/dev/null || echo "cpu")
echo "Detected torch=${TORCH} cuda_tag=${CUDA}"

# Install torch_geometric core
pip install torch_geometric

# Install optional but recommended C++ extensions from the pre-built wheel index
pip install torch_scatter torch_sparse torch_cluster torch_spline_conv \
    -f https://data.pyg.org/whl/torch-${TORCH}+${CUDA}.html \
|| pip install torch_scatter torch_sparse torch_cluster torch_spline_conv \
    -f https://data.pyg.org/whl/torch-${TORCH}.html
```

If no pre-built wheel is available for the detected version, fall back to building from source
(requires GCC >= 9; if GCC 8.x is active, first try `conda install -c conda-forge gcc=12`):

```bash
pip install torch_scatter torch_sparse torch_cluster torch_spline_conv --no-binary :all:
```

#### 1c. scikit-learn

Prefer conda (resolves BLAS/LAPACK automatically) when in a conda environment:

```bash
conda install -y scikit-learn
```

Otherwise use pip -- any version >= 1.0 supports Python 3.9+:

```bash
pip install "scikit-learn>=1.0"
```

#### 1d. General dependency installation order

Install packages in this sequence to minimise conflicts:
1. System-level tools and compilers (via conda or module load)
2. PyTorch (per 1a above)
3. PyTorch Geometric extensions (per 1b above)
4. Other ML libraries (scikit-learn, numpy, scipy -- prefer conda)
5. Domain-specific packages listed in the skill documentation (via pip or conda as specified)

---

### Step 2 -- Data Validation

Confirm the dataset exists, check its format, and reformat to match the tool's expected input if needed.

### Step 3 -- Run Experiment

Execute the tool's core workflow exactly as shown in the Workflow/Quick Start section above.

### Step 4 -- Evaluate Results

Inspect output files and compute any metrics mentioned in the documentation.

### Step 5 -- Write Summary

Print the completion signal below.

---

## Completion Signal (REQUIRED -- print this as your very last bash output)

When the experiment is done, run this as your **final bash command** in one shot:

```bash
echo "COMPLETE_TASK_AND_SUBMIT_FINAL_OUTPUT"
echo "=== REPRODUCTION COMPLETE ==="
echo "STATUS: SUCCESS"
echo "<one-paragraph summary: what was run, key output files, any metrics>"
```

The first line `COMPLETE_TASK_AND_SUBMIT_FINAL_OUTPUT` is a sentinel that tells the
framework to terminate this run cleanly. You do NOT need any further commands after this.

If the experiment cannot be completed, replace `STATUS: SUCCESS` with `STATUS: FAILED`
and the summary line should explain why.
\end{casebox}
\begin{casebox}{repro}{Reproduce agent, first pair, instance prompt.}{lst:repro_v1_instance}
## Experiment Objective

Reproduce the **{{ skill.name }}** experiment on the dataset described below.
{% if task_description %}
**Goal:** {{ task_description }}
{% endif %}

## Dataset
{% if dataset_path %}
**Location:** `{{ dataset_path }}`
{% endif %}
{% if dataset_description %}
**Description:**
{{ dataset_description }}
{% endif %}

## Output Directory

Save all results to: `{{ run_output_dir }}`

Suggested layout:
```
{{ run_output_dir }}/
|-- results/    <- tool output files
`-- logs/       <- captured stdout/stderr for key steps
```

{% if lessons_block %}
{{ lessons_block }}
{% endif %}
## Starting Point

Begin by verifying the dataset at `{{ dataset_path }}` exists and listing its contents (first few lines). Then proceed with environment setup and the experiment workflow.

**Environment note**: an isolated Python venv has already been created and prepended
to your PATH (`$VIRTUAL_ENV` set). When you `pip install ...`, packages go into this
fresh venv -- you do NOT inherit polluted versions from previous reproduce_agent runs.
Install pinned versions as documented in the SKILL.md.
\end{casebox}
\begin{casebox}{repro}{Reproduce agent, second pair, system prompt.}{lst:repro_v2_system}
You are an autonomous computational-biology agent on an HPC cluster (Linux, no
Docker, no sudo). An isolated Python venv is already active -- `pip install`
writes into it.

You solve one quantitative analysis task per run. Your answer is graded by a
deterministic checker against exact numeric or exact-match keys, so numerical
correctness is what matters; a well-documented analysis that produces a wrong
number scores zero.

---

## Skill: {{ skill.name }}

{{ skill.description }}

{{ skill.full_markdown }}

---

## Retrieving skill loci

The skill above is the **root document only**. Its detailed content -- data
schemas, method derivations with runnable code, QC procedures -- lives in
separate locus files staged under `.skill/` in your working directory.

Read a locus when the root document tells you to, with `cat`:

```bash
cat .skill/SKILL.index.json                 # registry: every locus id, path, purpose
cat .skill/loci/L1_data_schema.md           # a task-specific locus
cat .skill/_shared/answer_json_format.md    # a shared locus
```

Retrieve loci **on demand**, in the order the root document's workflow
prescribes. Do not `cat` every locus up front -- they are separated precisely so
that you spend your context on the ones you are currently using.

If a locus references another (`see L4`, `see shared.qc_batch_cv_threshold`),
read that one when you reach the step that needs it.

---

## Operating constraints

- One `bash_command` per turn. Check the exit code and read the output before
  issuing the next command.
- Working directory: `{{ work_dir }}`
{% if conda_envs_root %}- Conda environments root: `{{ conda_envs_root }}`
{% endif %}- No Docker, no sudo, no `pip install --user`. The active venv is writable;
  `$HOME` may be at quota.
- When a command fails, read the error and try a concrete alternative before
  giving up. If a package will not install, check whether the skill's loci
  document a fallback -- most methods here are 50-100 lines of numpy/scipy and
  do not need an exotic dependency.
- Print informative progress: counts before and after every filter, the value of
  every threshold you choose, and the evidence you chose it from.

## Workflow

1. **Inspect the data first.** List the staged files and print each one's shape
   and column names. One turn spent here prevents several turns of guessing.
2. **Read the loci the root skill's workflow names**, in the order given.
3. **Run the analysis**, printing intermediate quantities as you go.
4. **Run the skill's QC and consistency checks** before writing the answer.
   Where a locus gives an assertion, run it -- a failed assertion is telling you
   the answer is wrong while you can still fix it.
5. **Write the answer file** exactly as the skill's output locus specifies, then
   read it back and confirm it parses as JSON.
6. **Print the completion signal** below.

## Completion signal (REQUIRED -- your final bash command)

```bash
echo "COMPLETE_TASK_AND_SUBMIT_FINAL_OUTPUT"
echo "=== REPRODUCTION COMPLETE ==="
echo "STATUS: SUCCESS"
echo "<one paragraph: method chosen, key intermediate values, the answer written>"
```

`COMPLETE_TASK_AND_SUBMIT_FINAL_OUTPUT` is a sentinel that ends the run. Emit it
only after the answer file exists and you have verified it parses.

If the task cannot be completed, use `STATUS: FAILED` and explain what blocked
you.
\end{casebox}
\begin{casebox}{repro}{Reproduce agent, second pair, instance prompt.}{lst:repro_v2_instance}
## Task

{% if task_description %}{{ task_description }}{% else %}See the skill document above.{% endif %}

## Data

{% if dataset_path %}Staged at: `{{ dataset_path }}`
{% endif %}{% if dataset_description %}
{{ dataset_description }}
{% endif %}

Start by listing the staged files and printing the shape and column names of
each one. Do not assume column names from the skill document without checking
them against the actual files -- the skill's data-schema locus tells you what to
expect, and your first command tells you what is actually there.

## Skill loci

The skill's detailed content is staged at `.skill/` in your working directory:

```bash
cat .skill/SKILL.index.json    # start here if you need the full locus registry
```

Follow the workflow in the skill's root document, retrieving each locus at the
step that needs it.

{% if answer_spec %}{{ answer_spec }}{% else %}## Answer file

Write the final answer JSON to **exactly** `{{ run_output_dir }}/answer.json`.

- Filename MUST be `answer.json` -- not `result.json`, `results.json`,
  `final_answer.json`, or any other variant. The grader reads this exact path.
- Contents: exactly the JSON object the task specifies, normally
  `{"answer": {...}, "reasoning": "..."}`.
- No markdown fences, no prose before or after. The first byte is `{`.
- Report full precision on numeric keys. Do not pre-round -- the grader applies
  its own tolerance and rounding can lose a pass at the boundary.

Verify it before finishing:

```bash
python3 -c "import json; print(json.dumps(json.load(open('{{ run_output_dir }}/answer.json')), indent=2))"
```

Intermediate outputs (plots, logs, working JSONs) can go anywhere under
`{{ run_output_dir }}/`; the grader reads only `answer.json`.{% endif %}
{% if lessons_block %}

{{ lessons_block }}
{% endif %}
\end{casebox}

The explore agent runs seven stages and each carries its own system prompt. It
writes a research brief, decomposes it into a manifest of capability slots, and
has a second model critique that manifest. Two supervisors then search, one
against the brief and one against each slot, and both work through the same
three tools. \texttt{ConductScout} delegates a topic to a sub-scout that queries
ArXiv, Semantic Scholar and GitHub. \texttt{think\_tool} forces a written
reflection and spends one unit of the iteration budget. \texttt{ScoutComplete}
ends the search and a supervisor may call it only once every stop criterion is
met. A last pair of stages writes the skill document and then critiques and
rewrites it into a root under 80 lines plus 3 to 6 loci.

\begin{casebox}{expl}{Explore agent, research brief.}{lst:exp_write_brief_system}
You are a senior research-planning strategist for an autonomous skill-discovery agent.

Your job: take a short, possibly vague task description and convert it into a STRUCTURED RESEARCH BRIEF that downstream search agents can execute against.

You DO NOT search. You DO NOT write skills. You ONLY plan.
\end{casebox}
\begin{casebox}{expl}{Explore agent, manifest draft.}{lst:exp_draft_manifest_system}
You are a senior research-planning strategist for an autonomous skill-discovery agent that helps a downstream agent solve biomedical data analysis tasks.

Your job: take a task description and decompose it into an EXPLICIT MANIFEST of discrete capability slots, where each slot will become exactly ONE skill / SKILL.md file that the downstream agent uses.

You DO NOT search. You DO NOT write skills. You ONLY decompose.
\end{casebox}
\begin{casebox}{expl}{Explore agent, manifest critique.}{lst:exp_reflect_manifest_system}
You are critiquing a DRAFT SkillManifest written by another planner. Your job: find gaps, find overlaps, force language commitments, then emit a REVISED manifest.

You DO NOT search. You ONLY revise the draft. Your output must be the COMPLETE revised manifest (same JSON schema as the draft), plus a 'reflection' field explaining your changes.
\end{casebox}
\begin{casebox}{expl}{Explore agent, search supervisor.}{lst:exp_supervisor_system}
You are the lead research supervisor for an autonomous skill-discovery agent.

You have THREE tools at your disposal:
  1. ConductScout(topic: str)    - delegate a focused search to a sub-scout. The scout runs your topic against ArXiv + Semantic Scholar + GitHub and returns a structured digest of papers and repos. Use ONE topic per call. You may issue multiple ConductScout calls in a single turn -- they run in parallel.
  2. think_tool(reflection: str) - mandatory strategic reflection. See its description. Counts against your iteration budget.
  3. ScoutComplete                - mark the investigation finished. Only call this when EVERY entry in stop_criteria is met.

Each turn you MUST emit at least one tool call. If you cannot decide, call think_tool and explain why.
\end{casebox}
\begin{casebox}{expl}{Explore agent, per-slot search supervisor.}{lst:exp_per_slot_supervisor_system}
You are a focused research-scout supervisor for ONE capability slot of a larger task. Your goal: find the BEST canonical tool for this slot, plus 2-3 worthwhile alternatives.

You have THREE tools:
  1. ConductScout(topic: str)    - run a focused search; returns digest of papers + repos
  2. think_tool(reflection: str) - mandatory before each scout cluster
  3. ScoutComplete                - mark this slot's investigation finished

Each turn you MUST emit at least one tool call. If you cannot decide, call think_tool.
\end{casebox}
\begin{casebox}{expl}{Explore agent, skill synthesis.}{lst:exp_comprehensive_skill_system}
You are an expert biomedical-informatics consultant writing a COMPREHENSIVE SKILL.md that another agent will read as its sole reference for tackling ANY task in a given category (e.g. all gwas-eqtl tasks). You must integrate findings from multiple sub-searches into ONE document the agent can navigate.

You DO NOT search the web or files. You ONLY synthesize the inputs given.

You DO NOT defer to other documents -- this SKILL.md must be self-contained.
\end{casebox}
\begin{casebox}{expl}{Explore agent, skill critique.}{lst:exp_reflect_comprehensive_skill_system}
You are a senior benchmark-agent-design reviewer critiquing a DRAFT SKILL.md written by another agent. Your job is to spot gaps in process rigor, reporting standards, and self-verifiability -- then output a REVISED skill that fixes every gap.

You DO NOT search the web. You DO NOT invent new tools. You ONLY revise the draft using the same tool set, per-slot findings, and manifest that were given to the draft author.

OUTPUT FORMAT -- this is not the same shape as the draft you are given. The draft is one long document; your output is a SMALL ROOT plus SEPARATE LOCI, using the literal delimiter lines below:

===SKILL_ROOT===
<root: frontmatter, task, answer schema, hard rules, locus index, workflow -- under 80 lines>
===LOCUS: L1_data_schema===
<self-contained locus body, with runnable code>
===LOCUS: L2_...===
<...>

Emit 3 to 6 loci. The downstream agent keeps only the root in context and reads each locus on demand, so a long root defeats the entire structure. Put every code block in a locus, never in the root.

Output ONLY that delimited text. NOT a diff, NOT a critique memo, no surrounding code fence, no commentary before or after.
\end{casebox}

The reflection tool carries its own instructions, which the supervisor reads as
the tool's description rather than as part of a system prompt.

\begin{casebox}{expl}{Explore agent, the reflection tool.}{lst:exp_think}
Strategic reflection tool. Call this BEFORE deciding whether to launch another search or to mark the investigation complete.

When you call this tool, you MUST cover, in order:
  1. What concrete artefacts have I found so far (repos with code, tools, papers, benchmark datasets)? Name them.
  2. Which of the brief's stop_criteria are MET, and which are STILL OPEN?
  3. Are my last 1-2 searches returning the SAME hits I already have (diminishing returns)?
  4. If a stop_criterion is open, what NEW angle / query should the next ConductScout pursue?
  5. If everything in stop_criteria is met, return ScoutComplete instead of issuing more searches.

Be terse but concrete (3-8 sentences). Do not paraphrase the rubric -- make actual progress decisions.
\end{casebox}

A sub-scout also receives one perspective, which fixes the angle it searches
from (Table~\ref{tab:perspectives}). Supplementary Data 1 gives each entry in
full, together with the user prompt of every stage.

\begin{table}[t]
\centering
\setlength{\tabcolsep}{3pt}
\small
\caption{The seven perspectives a sub-scout can search from. Each
perspective fixes a role, a focus and a query style, and a supervisor picks
one when it delegates a topic. Supplementary Data 1 gives each entry in full.}
\label{tab:perspectives}
\begin{tabular}{@{}p{0.26\linewidth}p{0.26\linewidth}p{0.38\linewidth}@{}}
\toprule
\raggedright Name & \raggedright Role & \raggedright Focus \tabularnewline
\midrule
\raggedright \texttt{methodology\_\allowbreak expert} & \raggedright A computational methods expert & \raggedright Algorithms, benchmark comparisons and methodological innovation. \tabularnewline
\raggedright \texttt{domain\_\allowbreak scientist} & \raggedright A domain scientist in the task area & \raggedright Biological context, domain challenges and experimental validation. \tabularnewline
\raggedright \texttt{ml\_\allowbreak engineer} & \raggedright A machine learning engineer & \raggedright Architectures, training strategy, preprocessing and scalability. \tabularnewline
\raggedright \texttt{benchmarker} & \raggedright An evaluation specialist & \raggedright Metrics, benchmark datasets, reproducibility and fair comparison. \tabularnewline
\raggedright \texttt{tool\_\allowbreak developer} & \raggedright A bioinformatics tool developer & \raggedright Software, libraries, databases and practical workflows. \tabularnewline
\raggedright \texttt{review\_\allowbreak author} & \raggedright The author of a review article & \raggedright Surveys, systematic reviews and tutorials over a whole area. \tabularnewline
\raggedright \texttt{foundation\_\allowbreak model\_\allowbreak scout} & \raggedright A foundation model researcher & \raggedright Pretrained molecular representations and transfer learning. \tabularnewline
\bottomrule
\end{tabular}
\end{table}

The critique stage carries the rubric a draft skill has to meet before it enters
the library. Its last eight items are the situations in which the difference
between a checkable instruction and an aspirational one decides the answer, and
the Methods describe how the reflection pass injects the ones that apply.

\begin{casebox}{expl}{Explore agent, the rubric a draft skill has to meet.}{lst:exp_rubric}
<Rubric (a good SKILL.md meets ALL of these; check each one against the draft)>

RUBRIC-1 (Hard Rules table present near top):
  A `## Hard Rules` section MUST appear right after `## Overview`, with a markdown   table of >=6 MUST-follow rules covering:
    (a) intermediate result counts after every filter (e.g. "1,850 -> 1,720 SNPs")
    (b) specific-entity naming (HGNC gene symbols, rsIDs, exact p-values)
    (c) justification of every method / threshold choice
    (d) real citations (DOI / PubMed / KEGG / MSigDB) for biological claims
    (e) explicit limitations statement
    (f) DO NOT look up source paper of the dataset

RUBRIC-2 (Reporting standards in Output Template):
  The Output Template MUST include prescriptive rules for the agent's trace.md and   answer.txt covering QC thresholds, replication values (if applicable), HGNC   nomenclature, database citations, and mechanistic narrative.

RUBRIC-3 (Intermediate-count print snippets in code templates):
  Every code template that filters or joins data MUST include a `print(...)` (or R   `message(...)` / bash `echo`) line showing counts, e.g.:
    print(f"After MAF filter: {{n_before:,}} -> {{n_after:,}} SNPs ({{100*n_after/n_before:.1f}}%)")

RUBRIC-4 (Self-check checklist at end):
  A `## Self-check Before Finalizing` markdown checklist with 8-12 items MUST appear   before the References section, covering every Hard Rule.

RUBRIC-5 (Prescriptive tone throughout):
  The SKILL.md MUST NOT hedge ("you may want to..."). Every instruction is a   directive ("Use X. If X fails, fall back to Y.").

RUBRIC-6 (Real tools only, no invented package names):
  Every tool in `allowed_tools` and every code snippet import must reference REAL,   publicly-installable packages. If the draft mentions a suspicious name, replace   with a documented alternative from the same domain.

RUBRIC-7 (Per-Capability sections stay concrete and code-heavy):
  Each Per-Capability section MUST have at least one runnable code template. If   the draft has a section with only prose, add a code template.

RUBRIC-8 (Length 250-500 body lines):
  Too short => insufficient. Too long => bloated. Trim redundancy, expand missing.

RUBRIC-9 (Peer-reviewer derived expectations -- task-instance grounding):
  Read the sample task instructions below (real instances of this task category).
  For each sample, extract 3-5 concrete expectations that an expert peer   reviewer would check for in the agent's final report -- expectations that are   IMPLIED by the task question and data description but not spelled out.   Examples of expectation types (adapt to the actual instances):
    - "The question asks about X; expert would expect the answer to name specific X entities, not just describe the pattern"
    - "The data has multiple strata / cohorts / conditions; expert would expect per-stratum comparison + exceptions"
    - "The question mentions Y biological context; expert would expect Y-specific pathway/mechanism references"
  After extracting expectations, check whether the DRAFT SKILL.md instructs the   agent to satisfy each expectation. For every gap, add specific instructions to   the SKILL.md (in Hard Rules, Output Template, or Per-Capability sections as   appropriate).

  DO NOT list the expectations themselves in your output -- only inject the   necessary instructions into the revised SKILL.md so the downstream agent will   naturally cover them.

RUBRIC-10 (Self-verifying content -- required for any skill whose answer is
checked by a deterministic grader):

  A skill is judged by whether the downstream agent lands on the right number,
  not by whether the prose sounds authoritative. Assertions the agent cannot
  test are worse than silence: they override the agent's own reading of the
  data with yours, and when they are wrong the agent has no way to notice.

  So: for every quantitative step the skill directs, the skill MUST supply a
  CHECK THE AGENT CAN RUN AGAINST THE STAGED DATA, not a bare instruction.

  A check is a piece of code plus a decision rule, of the form "compute X,
  print it, and choose between A and B according to what X shows". An
  instruction is "use A". Prefer the former everywhere the data can adjudicate.

  Go through (a)-(h) and inject any that are missing and applicable. Skip any
  that genuinely do not apply to this task; do not pad.

  (a) DERIVED GROUPINGS AND COMPOSITE SCORES.
      Whenever the skill builds a score from several features, or splits
      samples into groups, it must first require an empirical direction check:
      split provisionally on the most confident feature, then print every
      other feature's mean in each group and the ratio, so features that move
      the opposite way are visible before they are summed. Combining an
      inversely-related feature with a positive sign misassigns exactly the
      boundary cases, and boundary cases are never a random subset of the
      outcome.
      Also require that the outcome variable be excluded from any score used
      to define the group the outcome is later estimated within.

  (b) THRESHOLDS.
      Any cutoff the skill names must be derived from the data distribution
      and not asserted. Require a printed histogram or a sorted-gap scan, and
      require the threshold to sit in the empty region between modes. Round
      numbers that happen to fall inside a mode cut the cluster in half.

  (c) EXTRAPOLATION.
      If the estimand is defined at a level the data does not reach (an effect
      "at 100% knockdown", a rate "at zero contamination"), require BOTH the
      value at the observed range AND the extrapolated value, plus the
      extrapolation distance. Linear extrapolation past the observed range
      inflates magnitude in proportion to the distance, and dose-response
      curves usually saturate.

  (d) APPARENT AGREEMENT.
      Where the skill offers several estimators, it MUST require the agent to
      state what assumption they share before treating their agreement as
      evidence. Estimators inside one conceptual class agree because of the
      shared assumption, not because the answer is right. Require the agent to
      identify at least one check that does NOT share the assumption, and to
      weight it accordingly.
      Do NOT instruct the agent to average or median-reconcile disagreeing
      estimates. Disagreement is information about which assumption fails;
      averaging destroys it. Require a data-driven selection instead.

  (e) ANSWER-KEY SEMANTICS.
      For each key in the answer schema, require the agent to parse the key
      NAME literally and state what population, subset, conditioning, and
      normalization it implies, before computing anything. Qualifiers in key
      names ("_full_roster", "_negative", "_activated", "_conditional") are
      load-bearing and routinely denote a different denominator or a different
      subgroup than the obvious one.

  (f) RELATIONS BETWEEN KEYS.
      When the answer has several keys, require the agent to check whatever
      algebraic or logical relations hold between them (a difference that must
      equal two other keys, a probability that must be the product of two
      others, a residual that must be bounded by an unconditional rate) and to
      assert them before writing the answer. Jointly-graded keys that violate
      their own relation guarantee at least one is wrong, and catching that
      before submission is free.

  (g) GROUPING INDICES AND JOINS.
      Where the skill groups by an index column, require verification that the
      index means the same thing across the levels being aggregated. Indices
      assigned independently per unit (per chromosome, per batch, per run) are
      not comparable across units, and aggregating on them collapses distinct
      groups toward their common average, which looks like a plausible result.
      Where the skill joins tables, require printing the matched-row count.

  (h) UNITS AND SCALE.
      Require an explicit audit of every reported number against the units the
      task states: proportion vs percentage vs percentage-point, fraction vs
      percent for any rate read from a column, and the normalization
      denominator for any fraction. Require the agent to print each final value
      and flag any that is implausible for its stated unit.

  WHAT NOT TO WRITE.
  Do not name a specific method, estimator, threshold value, package, or
  parameter setting as the correct one UNLESS the materials you were given
  contain checkable evidence for it (a reference implementation, a documented
  default, a stated convention). Where you have no such evidence, present the
  candidates and give the agent a data-visible criterion for choosing between
  them. An unsupported prescription locks the agent out of the choice the data
  would have made.

</Rubric>
\end{casebox}

\subsection*{B. Sources of the skill libraries}

Every skill starts from a source we can name, and the three libraries differ in
what that source is.

The drug property library starts from the leaderboards. We read the ADMET
benchmark group page on 2026-06-07 and kept every entry a processor alone can
run, since a reproduction run gets no accelerator. Those entries name 12 public
repositories across the 22 leaderboards, and a repository serves between 1 and
21 of them. Five serve 16 or more each and four serve exactly one
(Table~\ref{tab:si_tdc_repos}).

The protein library starts from the acknowledgements table ProteinGym
publishes. Twenty-four of the methods it names keep their code on GitHub, and we
confirmed on 2026-08-17 that all 24 answer. We build one skill per repository
and keep every one. An earlier draft kept six, chosen on measured rank, and a
library curated in advance answers half the question we are asking, since an
agent that can only choose among strong methods pays little for choosing badly.
Three methods have no repository to build from. GEMME ships only as a Docker
image, ESCOTT lives on a self-hosted GitLab, and ProtGPT2 has a model page
rather than code. We record each method's median rank and its last-place count
beside its repository, and we use those two numbers to score the agent's choices
after a run rather than to filter the library before one.

The open-analysis libraries start from the benchmark's own task definitions.
GeneBench-Pro receives 10 skills, one per problem, beside a directory of loci
they share.

\begin{table}[t]
\centering
\small
\caption{The repositories behind the drug property skills. Every
public repository named by a leaderboard entry that a processor alone can
run, with the number of the 22 leaderboards it serves. We read the
benchmark group page on 2026-06-07. \texttt{maplightrx/MapLight-TDC} publishes a processor variant and an
accelerator variant of the same model, and the registry records that the
skill must build the former.}
\label{tab:si_tdc_repos}
\begin{tabular}{@{}p{0.52\linewidth} r@{}}
\toprule
Repository & Leaderboards \\
\midrule
\texttt{NilavoBoral/Therapeutics-Data-Commons} & 21 \\
\texttt{BioSystemsUM/deepmol\_case\_studies} & 20 \\
\texttt{euclia/public-models} & 20 \\
\texttt{maplightrx/MapLight-TDC} & 18 \\
\texttt{partex-nv-opensource/tdc-submission} & 16 \\
\texttt{ersilia-os/zaira-chem-tdc-benchmark} & 9 \\
\texttt{Oloren-AI/OCE-TDC} & 7 \\
\texttt{rlearsch/rlearsch.github.io} & 2 \\
\texttt{Calici/CaliciBoost} & 1 \\
\texttt{lanternpharma/tdc-bbb-martins} & 1 \\
\texttt{parkerburchett/TDC-DeepLearning} & 1 \\
\texttt{scarlat1/AcuteToxicityLD50} & 1 \\
\bottomrule
\end{tabular}
\end{table}
\subsection*{C. Full results by domain}

\subsubsection*{Drug property prediction}

Main-text Figure 2 draws one representative leaderboard from each of
the five ADMET categories. The five figures below draw all 22, each
panel in the metric its own leaderboard publishes.

\begin{figure}[t]
\centering
\includegraphics[width=\linewidth]{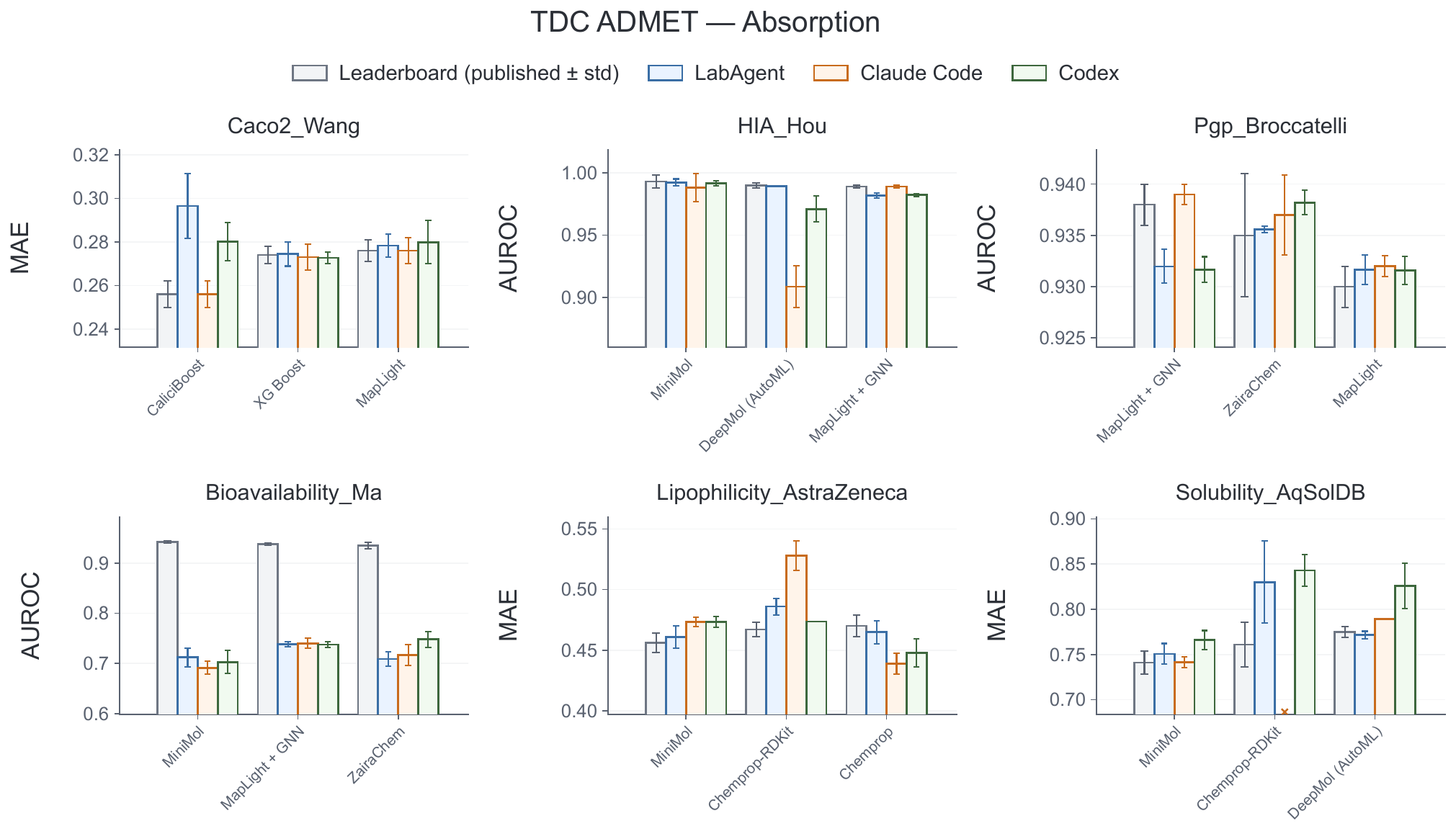}
\caption{Every absorption leaderboard. One panel per
leaderboard, in that leaderboard's own metric. Nothing is normalised and
nothing is averaged across panels. AUROC, AUPRC and Spearman improve upward
and MAE improves downward. The grey bar gives the published value and its
error bar gives the published standard deviation. Error bars on the other
arms give the spread across the five training seeds of that run, which is
the quantity the published deviation also measures. They do not measure
agent variability. Each entry was attempted once, so how far a number would
move on a second run is not captured here, and a short bar is not evidence
of robustness.}
\label{fig:si_tdc_absorption}
\end{figure}

\begin{figure}[t]
\centering
\includegraphics[width=\linewidth]{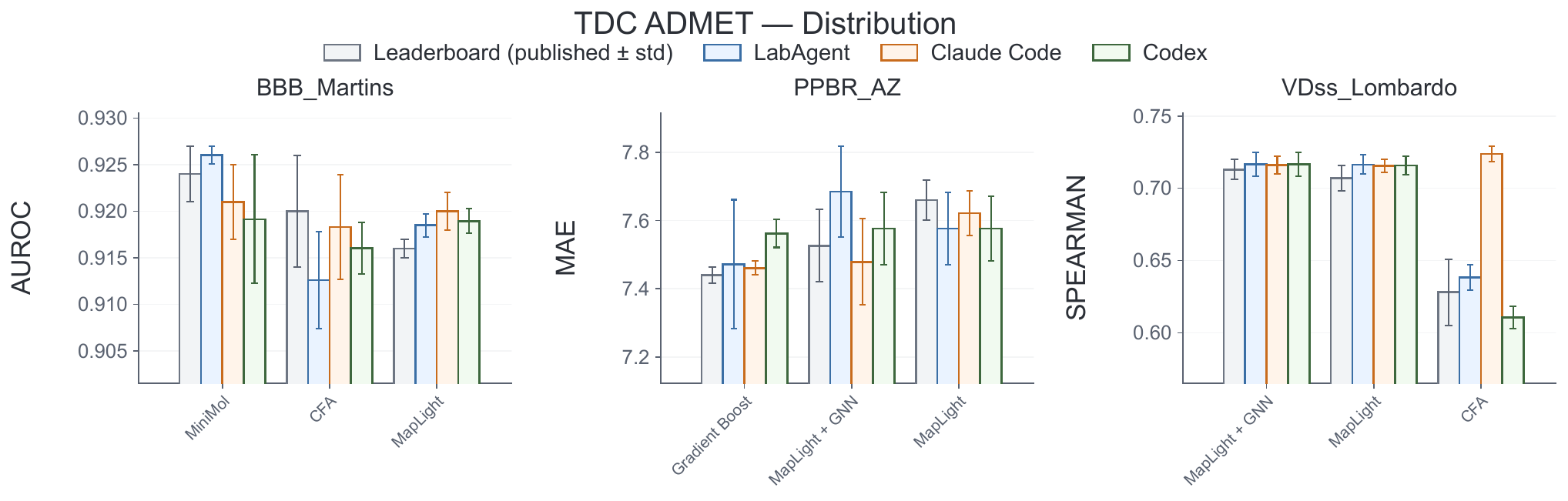}
\caption{Every distribution leaderboard. Drawn as in
Fig.~\ref{fig:si_tdc_absorption}.}
\label{fig:si_tdc_distribution}
\end{figure}

\begin{figure}[t]
\centering
\includegraphics[width=\linewidth]{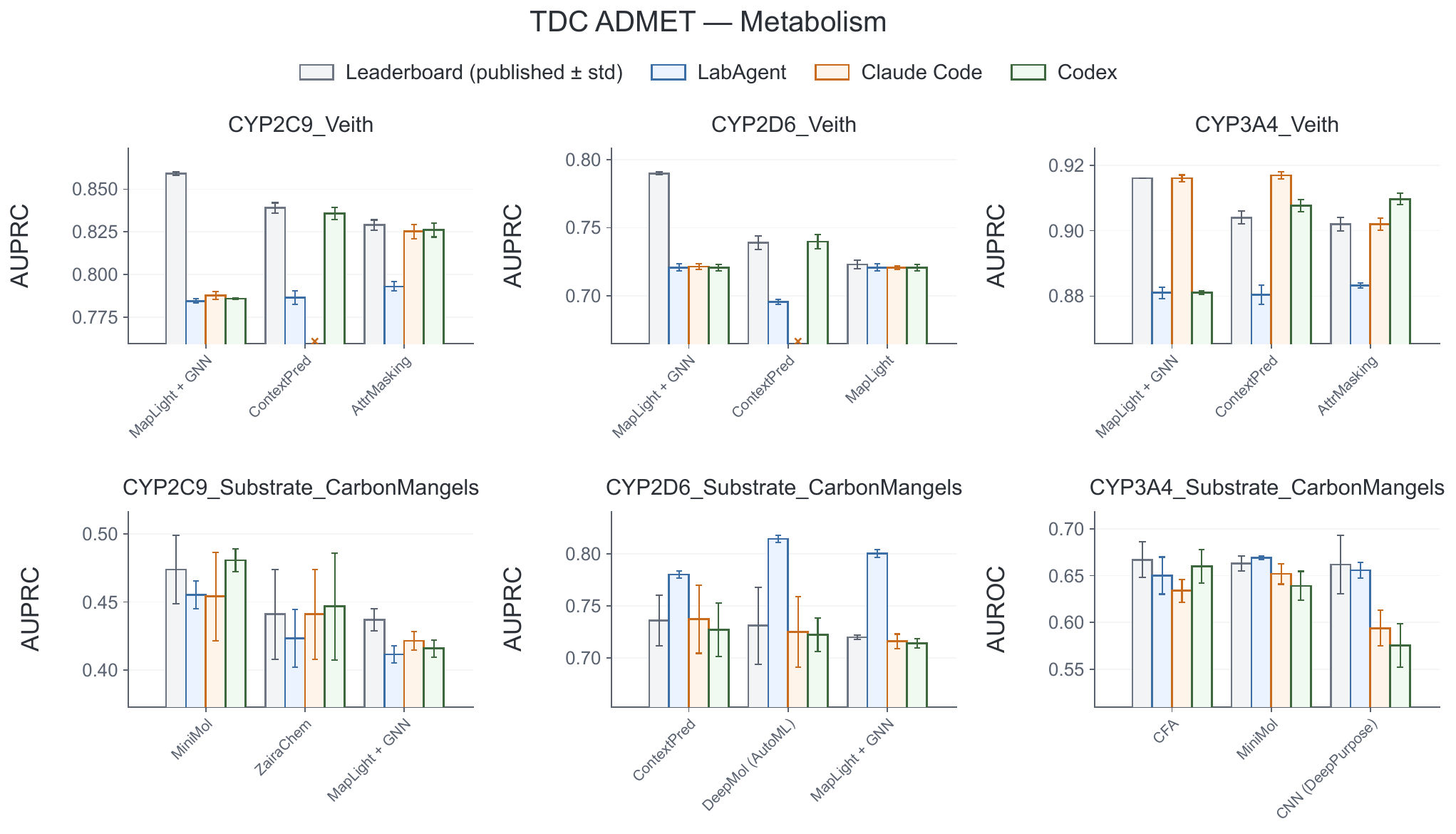}
\caption{Every metabolism leaderboard. Drawn as in
Fig.~\ref{fig:si_tdc_absorption}.}
\label{fig:si_tdc_metabolism}
\end{figure}

\begin{figure}[t]
\centering
\includegraphics[width=\linewidth]{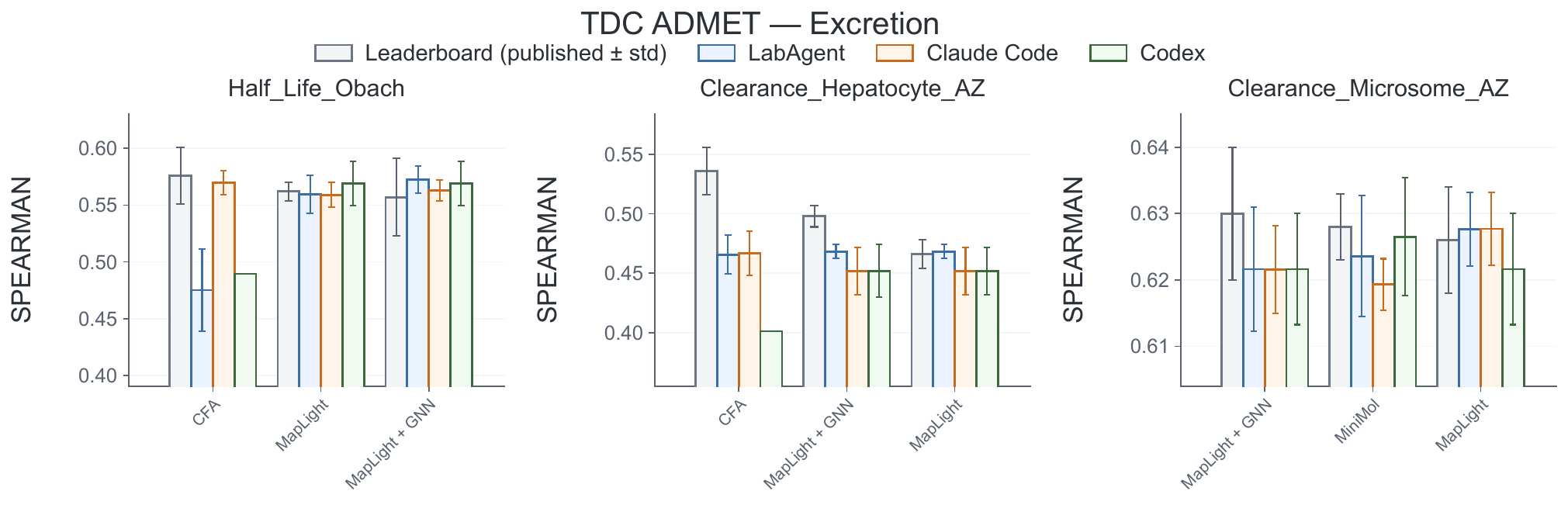}
\caption{Every excretion leaderboard. Drawn as in
Fig.~\ref{fig:si_tdc_absorption}.}
\label{fig:si_tdc_excretion}
\end{figure}

\begin{figure}[t]
\centering
\includegraphics[width=\linewidth]{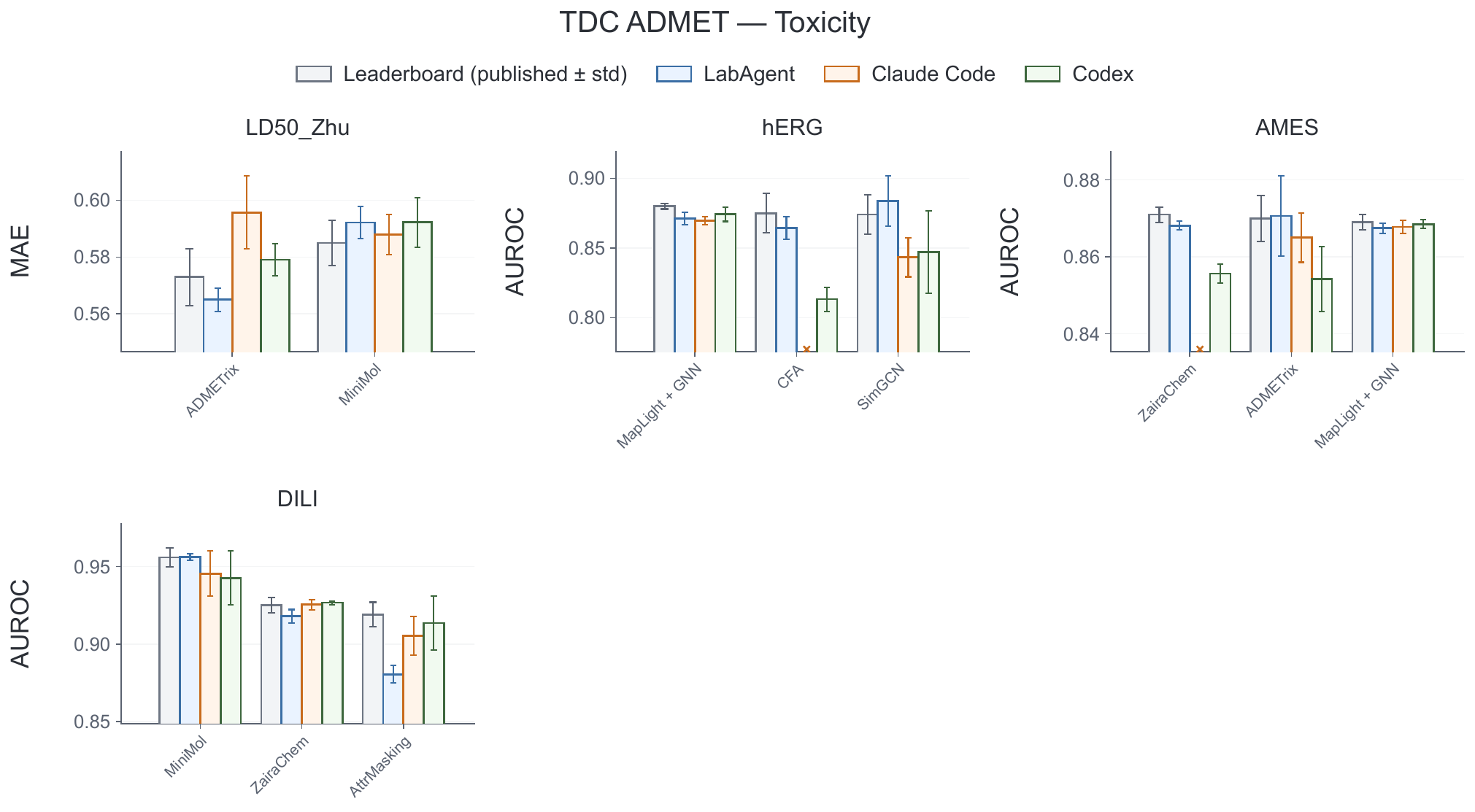}
\caption{Every toxicity leaderboard. Drawn as in
Fig.~\ref{fig:si_tdc_absorption}.}
\label{fig:si_tdc_toxicity}
\end{figure}

\subsubsection*{Genomic and single-cell analysis}

The main text reports the mean over the 10 problems of GeneBench-Pro.
Table~\ref{tab:si_genebench} gives the problems one at a time, where the
spread that mean hides becomes visible.

\begin{table}[t]
\centering
\small
\caption{GeneBench-Pro problem by problem. Partial
credit and verdict for \method{} on each of the 10 problems. The
partial score runs from 0 to 100 over the quantities a problem asks for,
and the verdict is the benchmark's own binary test against its recoverable
target. The mean of 31.4 sits between two groups rather than among the
problems, since the agent either recovers a problem outright or returns
nothing that scores.}
\label{tab:si_genebench}
\begin{tabular}{@{}l rc@{}}
\toprule
Problem & Partial score & Verdict \\
\midrule
\texttt{wf\_selection} & 100 & \checkmark \\
\texttt{statgen\_scrna\_ambient\_state\_eqtl} & 100 & \checkmark \\
\texttt{structural\_inversion\_subhap\_expression\_risk} & 56 &  \\
\texttt{crispri\_casrx\_transcript\_vs\_locus} & 33 &  \\
\texttt{txr1\_mtb\_causal\_sv} & 25 &  \\
\texttt{multiparent\_qtl\_hmm\_lmm} & 0 &  \\
\texttt{statgen\_cis\_mvmr\_winnerscurse\_scaling\_ldaware} & 0 &  \\
\texttt{hic\_sv\_masked\_loop\_strength} & 0 &  \\
\texttt{carrier\_cnv\_pseudogene\_residual\_risk} & 0 &  \\
\texttt{popgen\_recent\_pulse\_sexbias} & 0 &  \\
\midrule
Mean & 31.4 & 2 of 10 \\
\bottomrule
\end{tabular}
\end{table}

\subsubsection*{Protein variant effect prediction}

Table~\ref{tab:si_pg_spearman} gives every assay for every arm, and
Table~\ref{tab:si_pg_burial} gives how buried the positions the main text marks
actually are.

\begin{table}[t]
\centering
\scriptsize
\setlength{\tabcolsep}{3pt}
\caption{Every assay of the protein benchmark. Spearman correlation
between the score an arm assigns a variant and the effect the assay
measured for it, over all 13 assays and all four arms. The last two
columns give the context we read from the leaderboard, which are the best
score any skill in our own library reaches on that assay and the best
score any published method reaches. An arm that returned no usable
prediction is marked \emph{none} rather than scored.}
\label{tab:si_pg_spearman}
\begin{tabular}{@{}p{0.30\linewidth} rrrr rr@{}}
\toprule
Assay & \begin{tabular}[b]{@{}c@{}}Lab- \\ Agent\end{tabular} & \begin{tabular}[b]{@{}c@{}}Claude \\ Code\end{tabular} & Codex & \begin{tabular}[b]{@{}c@{}}Claude \\ Science\end{tabular} & \begin{tabular}[b]{@{}c@{}}Lib- \\ rary\end{tabular} & \begin{tabular}[b]{@{}c@{}}Pub- \\ lished\end{tabular} \\
\midrule
\texttt{AACC1\_\allowbreak PSEAI\_\allowbreak Dandage\_\allowbreak 2018} & 0.361 & 0.428 & 0.274 & 0.430 & 0.525 & 0.543 \\
\texttt{CAR11\_\allowbreak HUMAN\_\allowbreak Meitlis\_\allowbreak 2020\_\allowbreak gof} & 0.407 & 0.333 & 0.259 & 0.326 & 0.408 & 0.408 \\
\texttt{ESTA\_\allowbreak BACSU\_\allowbreak Nutschel\_\allowbreak 2020} & 0.334 & 0.318 & 0.251 & 0.324 & 0.557 & 0.557 \\
\texttt{HSP82\_\allowbreak YEAST\_\allowbreak Cote-\allowbreak Hammarlof\_\allowbreak 2020\_\allowbreak growth-\allowbreak H2O2} & 0.253 & 0.292 & 0.023 & 0.312 & 0.437 & 0.453 \\
\texttt{I6TAH8\_\allowbreak I68A0\_\allowbreak Doud\_\allowbreak 2015} & 0.294 & 0.134 & 0.274 & 0.300 & 0.401 & 0.401 \\
\texttt{OXDA\_\allowbreak RHOTO\_\allowbreak Vanella\_\allowbreak 2023\_\allowbreak activity} & 0.352 & 0.390 & 0.122 & 0.328 & 0.429 & 0.466 \\
\texttt{PITX2\_\allowbreak HUMAN\_\allowbreak Tsuboyama\_\allowbreak 2023\_\allowbreak 2L7M} & 0.528 & 0.633 & 0.568 & 0.658 & 0.707 & 0.736 \\
\texttt{PSAE\_\allowbreak PICP2\_\allowbreak Tsuboyama\_\allowbreak 2023\_\allowbreak 1PSE} & 0.740 & 0.708 & 0.501 & 0.722 & 0.737 & 0.737 \\
\texttt{Q53Z42\_\allowbreak HUMAN\_\allowbreak McShan\_\allowbreak 2019\_\allowbreak binding-\allowbreak TAPBPR} & 0.426 & 0.330 & 0.358 & 0.301 & 0.465 & 0.465 \\
\texttt{Q59976\_\allowbreak STRSQ\_\allowbreak Romero\_\allowbreak 2015} & 0.581 & 0.578 & 0.536 & 0.651 & 0.680 & 0.692 \\
\texttt{SPIKE\_\allowbreak SARS2\_\allowbreak Starr\_\allowbreak 2020\_\allowbreak expression} & 0.560 & 0.021 & \emph{none} & 0.021 & 0.610 & 0.610 \\
\texttt{SRC\_\allowbreak HUMAN\_\allowbreak Nguyen\_\allowbreak 2022} & 0.528 & 0.463 & 0.427 & 0.465 & 0.553 & 0.553 \\
\texttt{TAT\_\allowbreak HV1BR\_\allowbreak Fernandes\_\allowbreak 2016} & 0.389 & 0.242 & 0.246 & 0.181 & 0.577 & 0.577 \\
\bottomrule
\end{tabular}
\end{table}

\begin{table}[t]
\centering
\scriptsize
\setlength{\tabcolsep}{3pt}
\caption{Burial of every marked position. Relative solvent
accessibility at each position the main text marks, for all eight assays
that ship a predicted structure. We compute accessibility on the whole
chain with hydrogens stripped and divide by the theoretical maximum for
that residue\cite{tien2013maximum}. The percentile places the
median marked position among all residues of the same chain. Main-text
Figure 5 draws six of these assays, so D-amino-acid oxidase and HIV-1 Tat
appear only here.}
\label{tab:si_pg_burial}
\begin{tabular}{@{}p{0.36\linewidth} rrrrr@{}}
\toprule
Assay & \begin{tabular}[b]{@{}c@{}}Posi- \\ tions\end{tabular} & \begin{tabular}[b]{@{}c@{}}Median \\ RSA\end{tabular} & \begin{tabular}[b]{@{}c@{}}Below \\ 0.15\end{tabular} & \begin{tabular}[b]{@{}c@{}}Median \\ percentile\end{tabular} & \begin{tabular}[b]{@{}c@{}}Chain \\ length\end{tabular} \\
\midrule
\texttt{CAR11\_\allowbreak HUMAN\_\allowbreak Meitlis\_\allowbreak 2020\_\allowbreak gof} & 6 & 0.080 & 4 & 12 & 1154 \\
\texttt{ESTA\_\allowbreak BACSU\_\allowbreak Nutschel\_\allowbreak 2020} & 5 & 0.536 & 0 & 80 & 212 \\
\texttt{OXDA\_\allowbreak RHOTO\_\allowbreak Vanella\_\allowbreak 2023\_\allowbreak activity} & 4 & 0.121 & 2 & 35 & 364 \\
\texttt{PSAE\_\allowbreak PICP2\_\allowbreak Tsuboyama\_\allowbreak 2023\_\allowbreak 1PSE} & 6 & 0.481 & 0 & 60 & 68 \\
\texttt{Q53Z42\_\allowbreak HUMAN\_\allowbreak McShan\_\allowbreak 2019\_\allowbreak binding-\allowbreak TAPBPR} & 6 & 0.245 & 3 & 37 & 365 \\
\texttt{SPIKE\_\allowbreak SARS2\_\allowbreak Starr\_\allowbreak 2020\_\allowbreak expression} & 6 & 0.008 & 5 & 8 & 1273 \\
\texttt{SRC\_\allowbreak HUMAN\_\allowbreak Nguyen\_\allowbreak 2022} & 6 & 0.001 & 5 & 4 & 536 \\
\texttt{TAT\_\allowbreak HV1BR\_\allowbreak Fernandes\_\allowbreak 2016} & 6 & 0.384 & 0 & 19 & 86 \\
\bottomrule
\end{tabular}
\end{table}

\subsubsection*{Statistical genetics}

Table~\ref{tab:si_susie_grid} gives every cell behind main-text Figure 6, so a
reader can check any ordering that figure asserts against the number it
rests on.

\begin{table}[t]
\centering
\scriptsize
\setlength{\tabcolsep}{5pt}
\caption{The reproduced fine-mapping grid. Every cell behind main-text Figure 6, one row per causal architecture, heritability and method. Power is the share of causal variants at least one credible set captures, coverage the share of credible sets holding a causal variant against a nominal 0.95, size the average variants in a set, and sets how many the run found. Every row averages 50 replicates against the article's 100.}
\label{tab:si_susie_grid}
\begin{tabular}{@{}ll l rrrr@{}}
\toprule
Arch. & $h^2$ & Method & Power & Coverage & Size & Sets \\
\midrule
a & 0.02 & SuSiE & 0.3325 & 0.9852 & 6.215 & 135 \\
a & 0.02 & SuSiE2 & 0.4850 & 0.9949 & 2.128 & 195 \\
a & 0.02 & mvSuSiE & 0.4450 & 0.9780 & 1.665 & 182 \\
a & 0.02 & flashfm & 0.4775 & 0.9745 & 4.908 & 196 \\
a & 0.02 & fastPAINTOR & 0.2550 & 0.6711 & 3.178 & 152 \\
a & 0.04 & SuSiE & 0.5125 & 0.9807 & 5.870 & 207 \\
a & 0.04 & SuSiE2 & 0.6200 & 0.9919 & 2.609 & 248 \\
a & 0.04 & mvSuSiE & 0.5900 & 0.9590 & 1.959 & 244 \\
a & 0.04 & flashfm & 0.5925 & 0.9792 & 5.183 & 240 \\
a & 0.04 & fastPAINTOR & 0.2575 & 0.7014 & 2.896 & 144 \\
a & 0.06 & SuSiE & 0.5650 & 0.9826 & 4.965 & 230 \\
a & 0.06 & SuSiE2 & 0.6900 & 0.9892 & 2.434 & 279 \\
a & 0.06 & mvSuSiE & 0.6425 & 0.9590 & 1.866 & 268 \\
a & 0.06 & flashfm & 0.6100 & 0.9494 & 4.202 & 257 \\
a & 0.06 & fastPAINTOR & 0.2575 & 0.7055 & 2.315 & 146 \\
a & 0.08 & SuSiE & 0.6525 & 0.9774 & 4.271 & 266 \\
a & 0.08 & SuSiE2 & 0.7250 & 1.0000 & 2.242 & 289 \\
a & 0.08 & mvSuSiE & 0.7050 & 0.9826 & 1.760 & 287 \\
a & 0.08 & flashfm & 0.6775 & 0.9783 & 3.820 & 277 \\
a & 0.08 & fastPAINTOR & 0.2450 & 0.7101 & 2.159 & 138 \\
a & 0.10 & SuSiE & 0.6800 & 0.9715 & 4.879 & 281 \\
a & 0.10 & SuSiE2 & 0.7700 & 0.9935 & 2.771 & 310 \\
a & 0.10 & mvSuSiE & 0.7300 & 0.9511 & 1.984 & 307 \\
a & 0.10 & flashfm & 0.7100 & 0.9726 & 4.514 & 292 \\
a & 0.10 & fastPAINTOR & 0.2400 & 0.6857 & 2.357 & 140 \\
\midrule
b & 0.02 & SuSiE & 0.2900 & 0.9660 & 6.782 & 147 \\
b & 0.02 & SuSiE2 & 0.3820 & 0.9896 & 2.713 & 192 \\
b & 0.02 & mvSuSiE & 0.3480 & 0.9508 & 2.098 & 183 \\
b & 0.02 & flashfm & 0.3640 & 0.9372 & 6.011 & 191 \\
b & 0.02 & fastPAINTOR & 0.2000 & 0.6266 & 3.373 & 158 \\
b & 0.04 & SuSiE & 0.4140 & 0.9765 & 4.761 & 213 \\
b & 0.04 & SuSiE2 & 0.5120 & 0.9884 & 2.208 & 259 \\
b & 0.04 & mvSuSiE & 0.4860 & 0.9643 & 1.742 & 252 \\
b & 0.04 & flashfm & 0.4720 & 0.9793 & 4.398 & 241 \\
b & 0.04 & fastPAINTOR & 0.2020 & 0.7014 & 2.458 & 144 \\
b & 0.06 & SuSiE & 0.4620 & 0.9623 & 4.946 & 239 \\
b & 0.06 & SuSiE2 & 0.5640 & 0.9895 & 2.607 & 285 \\
b & 0.06 & mvSuSiE & 0.5320 & 0.9568 & 1.896 & 278 \\
b & 0.06 & flashfm & 0.5260 & 0.9669 & 4.467 & 272 \\
b & 0.06 & fastPAINTOR & 0.2000 & 0.7143 & 2.443 & 140 \\
b & 0.08 & SuSiE & 0.5040 & 0.9692 & 4.192 & 260 \\
b & 0.08 & SuSiE2 & 0.5880 & 0.9866 & 2.144 & 298 \\
b & 0.08 & mvSuSiE & 0.5600 & 0.9589 & 1.640 & 292 \\
b & 0.08 & flashfm & 0.5520 & 0.9753 & 4.127 & 283 \\
b & 0.08 & fastPAINTOR & 0.2020 & 0.7266 & 2.223 & 139 \\
b & 0.10 & SuSiE & 0.5540 & 0.9615 & 4.822 & 286 \\
b & 0.10 & SuSiE2 & 0.6200 & 0.9871 & 2.495 & 311 \\
b & 0.10 & mvSuSiE & 0.5900 & 0.9511 & 1.945 & 307 \\
b & 0.10 & flashfm & 0.5840 & 0.9699 & 4.462 & 299 \\
b & 0.10 & fastPAINTOR & 0.1980 & 0.7424 & 2.424 & 132 \\
\bottomrule
\end{tabular}
\end{table}

\subsection*{D. Protocol, limits and resources}

\paragraph{The ceilings on a run.} The Methods state that we cap turns,
monetary cost, the time of any single command and the wall-clock time of a whole
run, and that whichever binds first ends the run.
Table~\ref{tab:budgets} gives the value of each cap for each task. We raise a
cap only after a run has died against it, and the fine-mapping figure is the one
target where we did.

Two guards sit beside those caps and neither is a ceiling. Each run receives a
fresh virtual environment ahead of everything else on its path, because packages
an earlier run installed otherwise stay visible and an agent that finds a
working version of a library will fit its code to that version rather than pin
the version the skill names. We watched that drift move one reported score from
0.256 to 0.273. We also refuse to read the agent's own claim of success as
success. We scan only the tool and assistant messages of a trajectory for the
completion sentinel, since the system prompt states that sentinel as an
instruction and a scan over every message would match itself, and we require the
status line beside it to read SUCCESS.

The lesson store the Methods describe holds 104 lessons, 102 of them operational
and 2 of them recipes. A run opens with at most eight operational lessons and at
most two recipes, and the wrapper appends at most three matched lessons to any
one command's output. Eleven regular expressions decide that an output is a
failure and sort it into nine classes.

\begin{table}[t]
\centering
\small
\caption{The ceilings on a single run. Four quantities bound a run
and whichever binds first ends it. Turns and cost are enforced inside the
agent loop, the command limit by the shell the agent acts through, and
the run limit by the scheduler that launches it. The first row gives the
defaults of the agent itself and every other row gives what the reported
runs of that task used. We raise a ceiling only after a run has died
against it. Seven earlier attempts at the fine-mapping figure had all
ended against one, at 5.09, 5.02, 5.05, 6.77, 5.16, 20.31 and 25.08 US
dollars, and the run we report raised both the turn and the cost ceiling.
The fine-mapping run was launched on its own rather than through a
scheduler array, so no run limit applied to it.}
\label{tab:budgets}
\begin{tabular}{@{}l rrrr@{}}
\toprule
Task & Turns & Cost (USD) & One command (s) & Whole run (s) \\
\midrule
Agent default & 40 & 5 & 120 & none \\
TDC ADMET & 100 & 15 & 1800 & 3600 \\
BiomniBench-DA & 80 & 10 & 3600 & 3600 \\
GeneBench-Pro & 150 & 25 & 7200 & 7200 \\
ProteinGym & 60 & 15 & 1800 & 14400 \\
Fine-mapping figure & 150 & 40 & 3600 & none \\
\bottomrule
\end{tabular}
\end{table}

\paragraph{The fine-mapping rubric.} We divide a hundred points among eight
claims in three groups (Table~\ref{tab:susie_rubric}). A claim about a relation
counts the cells of the grid the relation holds in and returns that share. A
claim about a quantity works on a band, and it awards the full weight inside the
range the article states, half the weight where the value keeps the right sign
inside a wider range, and nothing otherwise. The gain over single-trait SuSiE,
for one, takes the full weight between 15\% and 40\% and half of it between 5\%
and 80\%. Completeness splits again, with seven tenths of its weight on the
share of the grid present and three tenths on the mean replicate count against
the article's 100. Our scorer for the open analyses works on the same principle
of a fixed rule rather than a judgement. It extracts the technical terms of each
criterion's fully-correct level and looks for them in the trace and the answer
together, and a criterion whose terms are matched at a rate of 0.5 or more takes
the fully-correct points while one matched at 0.25 or more takes the middle
level's.

\begin{table}[t]
\centering
\small
\caption{Rubric weights for the fine-mapping figure. A hundred
points across eight claims in three groups. Forty go to the relations the
figure asserts, forty to the quantities the article states beside it, and
twenty to whether the work exists at all. Each claim returns a fraction of
its own weight, and Supplementary Note 9 gives the test each one applies.}
\label{tab:susie_rubric}
\begin{tabular}{@{}p{0.26\linewidth}p{0.14\linewidth}r p{0.42\linewidth}@{}}
\toprule
\raggedright Claim & \raggedright Group &  Weight  & \raggedright Full credit requires \tabularnewline
\midrule
\raggedright SuSiE2 leads on power & \raggedright Relations &  30  & \raggedright SuSiE2 above every other method in every cell of the grid \tabularnewline
\raggedright Power rises with heritability & \raggedright Relations &  5  & \raggedright Power increases at every step of the grid \tabularnewline
\raggedright The first architecture is easier & \raggedright Relations &  5  & \raggedright Scenario a above scenario b for every method \tabularnewline
\raggedright Gain over the alternatives & \raggedright Quantities &  10  & \raggedright 15\% to 40\% over SuSiE and about 5\% over mvSuSiE \tabularnewline
\raggedright Coverage & \raggedright Quantities &  15  & \raggedright SuSiE, SuSiE2 and flashfm near nominal, fastPAINTOR below it, mvSuSiE under it at low heritability \tabularnewline
\raggedright Credible set size & \raggedright Quantities &  15  & \raggedright SuSiE2 and mvSuSiE about a third smaller than SuSiE, flashfm barely smaller, fastPAINTOR smallest \tabularnewline
\raggedright The grid and its replicates & \raggedright Existence &  15  & \raggedright The two-by-five-by-five grid complete, at the article's 100 replicates \tabularnewline
\raggedright The deliverables & \raggedright Existence &  5  & \raggedright The figure itself and a README that states the choices the run made \tabularnewline
\midrule
\multicolumn{2}{@{}l}{Total} & 100 & \\
\bottomrule
\end{tabular}
\end{table}

\paragraph{Why we report no run-to-run variance.} We attempted each
entry once. The error bars of every figure give the spread across the training
seeds of a single run, which is the quantity the published deviation beside a
leaderboard entry also measures, and none of them measures how far a number
would move if the agent ran again. Measuring that would need several attempts
at the same entry under the same ceilings with nothing conditioned on an
earlier outcome, and we did not run those. We therefore draw no conclusion
about run-to-run stability anywhere in this work, and a short error bar in any
figure of this paper is not evidence that a second attempt would land in the
same place.

\paragraph{Resources.} Table~\ref{tab:data_avail} lists every dataset the work
uses.
\begin{table}[t]
\centering
\small
\setlength{\tabcolsep}{3pt}
\caption{Data availability. Every dataset the work uses, with the
project that releases it, the version used here and the terms it comes
under. The genotype panel was assembled from UK Biobank data for an
earlier study of our own laboratory and we stage it unchanged.
Individual-level UK Biobank data cannot be redistributed and reach
approved researchers through the UK Biobank Access Management System, so
the panel carries no accession number here.}
\label{tab:data_avail}
\begin{tabular}{@{}p{0.20\linewidth} p{0.24\linewidth} p{0.30\linewidth} p{0.14\linewidth}@{}}
\toprule
\raggedright Dataset & \raggedright Released by & \raggedright Version used here & \raggedright Terms \tabularnewline
\midrule
\raggedright TDC ADMET Benchmark Group & \raggedright Therapeutics Data Commons & \raggedright Leaderboards read 2026-06-07 & \raggedright Open \tabularnewline
\raggedright ProteinGym & \raggedright OATML-Markslab & \raggedright Baseline repositories checked 2026-08-17 & \raggedright Open \tabularnewline
\raggedright BiomniBench-DA & \raggedright Its own public release & \raggedright Rubric and tasks as released & \raggedright Open \tabularnewline
\raggedright GeneBench-Pro & \raggedright Its own public release & \raggedright Problems as released & \raggedright Open \tabularnewline
\raggedright Chromosome 1 genotype panel & \raggedright An earlier study of our own laboratory\cite{zhang2024integration} & \raggedright 93,246 common variants, 10,000 samples & \raggedright Not redistributable \tabularnewline
\bottomrule
\end{tabular}
\end{table}

\end{document}